\pdfoutput=1
\documentclass[12pt,amsmath,amssymb,longbibliography,nofootinbib,showkeys,eprint]{revtex4-2}

\usepackage[activate={true,nocompatibility},final,tracking=true,kerning=true,spacing=true]{microtype}
\usepackage{epigraph}
\usepackage{graphicx}
\usepackage{epstopdf}
\usepackage{braket}
\usepackage{bm}
\usepackage{numprint}
\usepackage{float}
\usepackage[T2A,T1]{fontenc}
\usepackage[utf8]{inputenc}
\usepackage{hhline}
\usepackage{xfrac}
\usepackage{amsthm}
\usepackage{makecell}
\usepackage{dsfont}
\usepackage{calc}
\usepackage{seqsplit}
\usepackage{circuitikz}
\usepackage{hyperref}
\hypersetup{
    colorlinks=true
}
\usepackage{orcidlink}

\begin{document}
% $Id: QuantumChannelLearning.tex,v 1.404 2025/01/03 17:31:18 mal Exp $
\preprint{V.M.}
\title{Quantum Channel Learning}

\author{Mikhail Gennadievich \surname{Belov}}
\email{mikhail.belov@tafs.pro}
\affiliation{Lomonosov Moscow State University,  Faculty of Mechanics and Mathematics,
   GSP-1,  Moscow, Vorob'evy Gory, 119991, Russia}

\author{Victor Victorovich \surname{Dubov}}
\email{dubov@spbstu.ru}
\affiliation{Peter the Great St. Petersburg Polytechnic University, 195251, Russia}

\author{Alexey Vladimirovich \surname{Filimonov}\,\orcidlink{0000-0002-2793-5717}}
\email{filimonov@rphf.spbstu.ru}
\affiliation{Peter the Great St. Petersburg Polytechnic University, 195251, Russia}

\author{Vladislav Gennadievich \surname{Malyshkin}\,\orcidlink{0000-0003-0429-3456}} 
\email{malyshki@ton.ioffe.ru}
\affiliation{Ioffe Institute, Politekhnicheskaya 26, St Petersburg, 194021, Russia}

\date{June, 6, 2024}

\begin{abstract}
\baselineskip16pt 
\begin{verbatim}
$Id: QuantumChannelLearning.tex,v 1.404 2025/01/03 17:31:18 mal Exp $
\end{verbatim}
The problem of an optimal mapping 
between Hilbert spaces
\emph{IN} 
and
\emph{OUT}, 
based
on a series of density matrix mapping measurements
$\rho^{(l)} \to \varrho^{(l)}$, $l=1\dots M$,
is formulated as an optimization problem maximizing the total fidelity
$\mathcal{F}=\sum_{l=1}^{M} \omega^{(l)}
F\left(\varrho^{(l)},\sum_s B_s \rho^{(l)} B^{\dagger}_s\right)$
subject to probability preservation constraints
on  Kraus operators $B_s$.
For $F(\varrho,\sigma)$ in the form that
total fidelity can be represented
as a quadratic form with superoperator
$\mathcal{F}=\sum_s\left\langle B_s\middle|S\middle| B_s \right\rangle$
(either exactly or as an approximation)
an iterative algorithm is developed.
The work introduces 
two important generalizations of unitary learning:
1. \emph{IN}/\emph{OUT} states are represented as density matrices.
2. The mapping itself is formulated as a mixed unitary quantum channel
$A^{\emph{OUT}}=\sum_s |w_s|^2 \mathcal{U}_s A^{\emph{IN}} \mathcal{U}_s^{\dagger}$
(no general quantum channel yet).
This marks a crucial advancement from the commonly studied unitary mapping of pure states
$\phi_l=\mathcal{U} \psi_l$
to a quantum channel,
what allows us
to distinguish
probabilistic mixture of states and their superposition.
An application of the approach is  demonstrated
on unitary learning of density matrix mapping
$\varrho^{(l)}=\mathcal{U} \rho^{(l)} \mathcal{U}^{\dagger}$,
in this case a quadratic on $\mathcal{U}$ fidelity can be
constructed by considering $\sqrt{\rho^{(l)}} \to \sqrt{\varrho^{(l)}}$
mapping, and on a quantum channel,
where quadratic on $B_s$  fidelity is an approximation ---
a quantum channel is then obtained as a hierarchy of unitary mappings,
a mixed unitary channel.
The approach can be applied to studying quantum inverse problems,
variational quantum algorithms, quantum tomography, and more.
A software product implementing the algorithm
\href{http://www.ioffe.ru/LNEPS/malyshkin/code_polynomials_quadratures.zip}{is available}
from the authors.
\end{abstract}

\maketitle
\noindent DOI: \href{https://doi.org/10.1103/PhysRevE.111.015302}{10.1103/PhysRevE.111.015302}
\newpage

{\noindent Dedicated to the memory of Ivan Anatol'evich Komarchev}

\section{\label{intro}Introduction}
The form of knowledge representation stands out as a crucial and distinctive feature in any approach to Machine Learning (ML).
The utilization of unitary operators
for knowledge representation
has garnered increasing attention recently\cite{bisio2010optimal,arjovsky2016unitary,hyland2017learning}.
In nature most of dynamic equations
are equivalent to a sequence of infinitesimal unitary transformations:
Newton, Maxwell, Schr\"{o}dinger equations.
This inherent connection with unitary transformations
makes it particularly appealing to represent knowledge in this form. 
Most of the existing works consider algorithms that take wavefunction
as input and construct a unitary operator
providing a high value of mapping
fidelity\cite{arjovsky2016unitary,hyland2017learning,lloyd2019efficient};
they are mostly different in
parametrization of unitary operator
\cite{RazaUnitaryParametrization} and optimization details.
Such a pure state to pure state mapping
is a limited form of quantum evolution.
There are a few works\cite{pai2019matrix,huang2021learning,gonzalez2022learning,yu2023optimal}
that consider unitary learning with density matrix input.
Real systems probability should be described by a mixed state
(density matrix)
what allows
to distinguish
probabilistic mixture of states and their superposition.
There are two sources of mixed states:
1. The input data itself can be in mixed state.
2. $\emph{IN}\to\emph{OUT}$ mapping of a general quantum channel
can transform a pure state input into a mixed state output. Unitary mapping
is a simple example of a quantum channel converting
a pure state into a pure state.

In our previous works\cite{malyshkin2022machine,belov2024partiallyPRE}
the quantum mechanics inverse problem of optimal unitary mapping of pure states was converted 
to a \href{https://en.wikipedia.org/wiki/Quadratically_constrained_quadratic_program}{QCQP} (Quadratically Constrained Quadratic Program) problem,
a novel algebraic problem (\ref{eigenvaluesLikeProblem}) was formulated,
and an efficient iterative global optimization algorithm was developed.
This algorithm can be applied to quantum inverse problems,
variational quantum algorithms \cite{cerezo2021variational,park2024hamiltonian,wang2024variationalPRL} with a cost function in the form
$C(\theta)= \mathrm{Tr}\, O \mathcal{U}(\theta)\rho_0\mathcal{U}^{\dagger}(\theta)$,
quantum tomography\cite{MAURODARIANO2003205,lvovsky2009continuous,torlai2023quantum,ahmed2023gradient},
various classical problems, and many others.
We do not have a formal proof of the developed algorithm's convergence,
but among the millions of test runs,
only a few did not converge to the global maximum. This could possibly be caused by numerical instability.
In the current work, we generalize our algorithm to mixed states and further extend it to quantum channels.
This generalization is possible as long as the total fidelity can be represented
as a quadratic form on quantum channel mapping operators.

In the case of unitary mapping ($D=n$, $N_s=1$) of mixed states
there are several good choices for fidelity, for example
a quantum channel mapping the square root of
density matrix
$\sqrt{\rho}\to\sqrt{\varrho}$
with a unitary operator (\ref{operatorTransform}),
it provides the exact fidelity as a quadratic form.
Together with the available iterative algorithm,
this problem may be considered mostly solved.

In the case of a general quantum channel (\ref{KrausOperator})
constructing fidelity as the quadratic form 
of Eq. (\ref{FasOperatorsProduct})
on quantum channel mapping operators can be challenging.
We have developed several approximations that can be used with the iterative algorithm and have constructed
a hierarchy of mapping operators (\ref{FexpansionOrt})
as a mixed unitary channel.
A demonstration of several approximations
is presented in Section \ref{DemoHierarchy}.
The problem of finding a good quadratic (on mapping operators) fidelity
for Kraus rank $N_s>1$ quantum channels
requires more research and is related
to a physical meaning of quantum channel mapping.
Even when fidelity is expressed as the quadratic form of Eq. (\ref{FasOperatorsProduct}),
we currently only have a numerical algorithm to approximate the data using a mixed unitary channel.
A general reconstruction of quantum channels is not yet available.

The paper is organized as follows. After formulating the problem,
the unitary mapping of mixed states is considered in Section \ref{UnitaryMappingOfMixedStates}.
Section \ref{Kraushierarchy} focuses on the construction of a hierarchy of unitary operators.
In the conclusion, we discuss the results obtained for the problem of quantum channel reconstruction.
Appendix \ref{numericalSolutionSect} provides a detailed description of the numerical solution that uses algebraic techniques.
Appendix \ref{SchodingerNonStationary} presents a generalization of the novel algebraic problem (\ref{eigenvaluesLikeProblem})
to the nonstationary case and introduces a time-dependent Schr\"{o}dinger-like equation
for unitary operator dynamics (\ref{SchodingerNonStationaryEq}).
Appendix \ref{StatesWithMemory} explores potential generalizations to states with memory,
while Appendix \ref{compComplexityEstimation} estimates the algorithm's computational complexity.

This paper is accompanied by a software which
\href{http://www.ioffe.ru/LNEPS/malyshkin/code_polynomials_quadratures.zip}{is available}
from Ref. \cite{polynomialcode};
all references to code in the paper correspond to this software.

\section{\label{formulationOfTheProblem}Formulation of the Problem}
In \cite{belov2024partiallyPRE}, we considered data of vector-to-vector pure state mapping,
 $l=1\dots M$, 
\begin{align}
\psi_l(\mathbf{x})& \to \phi_l(\mathbf{f})
  & \text{weight $\omega^{(l)}$}  \label{mlproblemVector}
\end{align}
The weights $\omega^{(l)}$ are typically all equal to 1.\footnote{
\label{normalizingToNumberObservations}
Some authors use $\omega^{(l)} = 1/M$ to normalize the fidelity to the range $[0: 1]$.
However, this approach is inconvenient because $\mathcal{F}$ is no longer an
\href{https://en.wikipedia.org/wiki/Intensive_and_extensive_properties}{extensive quantity}
in the sense that, for two sets of observations, the total fidelity is no longer the sum of the two --- losing its additive property.
For this reason, normalizing to the number of observations is preferred over normalizing to $[0: 1]$.
}
However, they can be set to different values if the observations
are made with different accuracy,
this is particularly convenient for classical systems.
The usefulness of  $\omega^{(l)}$ weights also arises in models that comprise
a number of identical mappings;
in this case one may consider only distinct mappings and set
$\omega^{(l)}$
equal to the number of times the observation was encountered in the sample.
The problem of maximizing
total fidelity
\begin{align}
\mathcal{F}&
=\sum\limits_{l=1}^{M} \omega^{(l)}
\Big|\Braket{\phi_l|\mathcal{U}|\psi_l}\Big|^2
\xrightarrow[{\mathcal{U}}]{\quad }\max
\label{allProjUKxfAppendix}
\end{align}
was considered, and a numerical algorithm finding the
global maximum of $\mathcal{F}$ was developed.
Found solution $\mathcal{U}$,
a partially unitary matrix of $\dim(\emph{OUT})\times \dim(\emph{IN})$ ($D\times n$)
\begin{align}
A^{\emph{OUT}}&=\mathcal{U} A^{\emph{IN}} \mathcal{U}^{\dagger}
\label{operatorTransform}
\end{align}
converts
any operator between two Hilbert spaces,
for example it converts a pure state $A^{\emph{IN}}=\Ket{\psi}\Bra{\psi}$
into a pure state $A^{\emph{OUT}}=\Ket{\mathcal{U}|\psi}\Bra{\psi|\mathcal{U}^{\dagger}}$.
For $D=n$ it is a
trace preserving map,
and for $D<n$ it is a 
\href{https://en.wikipedia.org/wiki/Quantum_channel}{trace decreasing map}
quantum channel.
The input data (\ref{mlproblemVector})
represent pure state to pure state mapping. 
This type of mapping is not the most
general, it cannot describe
systems with effects like
\href{https://en.wikipedia.org/wiki/Quantum_decoherence}{quantum decoherence}. For example a simple
\href{https://en.wikipedia.org/wiki/Markov_chain}{Markov chain}
system converts a pure state $\Ket{\psi}\Bra{\psi}$
into a density matrix state, see Appendix I of Ref. \cite{malyshkin2019radonnikodym}.

In this work a more general mapping
between Hilbert spaces
\emph{IN} of
$\Ket{ \psi}$ (dimension $n$)
and
\emph{OUT} of
$\Ket{\phi}$ (dimension $D$)
is considered.
\href{https://en.wikipedia.org/wiki/Quantum_operation#Kraus_operators}{Kraus' theorem}
determines the most general form
of mapping between Hilbert spaces\cite{jamiolkowski1972linear,kraus1983states,joos2013decoherence}:
\begin{align}
  A^{\emph{OUT}}&=\sum\limits_{s=0}^{N_s-1} B_s A^{\emph{IN}} B^{\dagger}_s
  \label{KrausOperator}
\end{align}
the number of terms in the sum $N_s$ is called the Kraus rank,
see Choi's theorem \cite{choi1975completely}
and Belavkin's Radon-Nikodym theorem for completely positive maps \cite{belavkin1986radon}.
The Kraus operators $B_s$ satisfy 
the constraints that unit $A^{\emph{IN}}$
is converted to unit $A^{\emph{OUT}}$. There are other options
to construct a quantum channel, in \cite{malyshkin2022machine,belov2024partiallyPRE}
we considered a quantum channel
that transforms
the Gram matrix
in space $\Ket{\psi}$ into the Gram matrix in space $\Ket{\phi}$;
with proper regularization it can be reduced to the same form.
\begin{align}
  \sum\limits_{s=0}^{N_s-1} B_sB_s^{\dagger}&=\mathds{1} \label{constraintKrauss}
\end{align}
We further generalize Kraus operators $B_s$ by considering them to be rectangular
$D\times n$ matrices
instead of being limited to Hilbert space mapping into itself.
The term ``partially Kraus'' is used in the same sense
as ``partially unitary'' --- we consider rectangular
matrices $D\le n$ mapping Hilbert spaces of different dimensions.\footnote{
\label{condTraceFootnote}
Here, instead of the usual constraint
$\sum_{s=0}^{N_s-1} B_s^{\dagger}B_s=\mathds{1}$ (\ref{constraintKraussSpur})
we use the constraint
(\ref{constraintKrauss})
to fully utilize our numerical algorithm in the case of $N_s=1$.
Constraint (\ref{constraintKrauss}) imposes $D(D+1)/2$ independent conditions,
whereas constraint (\ref{constraintKraussSpur}) imposes $n(n+1)/2$ independent conditions.
When $D=n$ both definitions coincide. However, for $N_s=1$ and $D<n$,
the constraint (\ref{constraintKraussSpur}) fails to satisfy the minimal Kraus rank condition (\ref{minKrausRank}),
but the constraints (\ref{constraintKrauss}) are applicable in the numerical study of trace-decreasing maps when $N_s=1$ and $D<n$.
Additionally, the Lagrange multiplier calculation method described in Appendix \ref{lagrangeMultipliersCalculation},
along with the convergence-helper constraints in Appendix \ref{LinearConstraints},
facilitates the incorporation of quadratic form constraints of any type.
}
For $D<n$ a partially unitary quantum channel (\ref{operatorTransform})
does not preserve the density matrix trace. For a large enough $N_s$, as in the condition (\ref{minKrausRank}), the form
of Eq. (\ref{KrausOperator}) allows us to construct a trace preserving quantum
channel, for example to implement an operation
of
\href{https://en.wikipedia.org/wiki/Partial_trace}{partial trace}.

Kraus operators selection is not unique.
In addition to the familiar $\exp(i\varphi)$ phase factor ($\pm 1$ for real space),
there is a
\href{https://en.wikipedia.org/wiki/Gauge_theory}{gauge}
that regulates redundant degrees of freedom.
The most well-known is the canonical form:
\begin{align}
\mathrm{Tr}B_sB^{\dagger}_t&=0 &\text{for $s\ne t$}
\label{KrausCanonicForm}
\end{align}
However, other gauges can be used.
See Appendix \ref{KrausCanonicMethod} for an algorithm
transforming Kraus operators to canonical form.

Back in \cite{malyshkin2019radonnikodym,malyshkin2022machine}
we considered, instead of (\ref{mlproblemVector}),
the data of density matrix to density matrix mapping (\ref{mlproblemVectorDensityMatrix}) ---
just replace $S_{jk;j^{\prime}k^{\prime}}$ from (\ref{SfunctionalClassic}) by (\ref{StendorDensityMatrix}) or (\ref{StendorDensityMatrixSQRTrho}).
Whereas input/output data in density matrix form
distinguishes
probabilistic mixture of states and their superposition,
the transform (\ref{operatorTransform}) cannot map such data exactly,
it has insufficient expressive power. 
Only a general quantum channel  (\ref{KrausOperator})  mapping
has sufficient expressive power to construct
learning models from general data,
for example with a decoherence effect,
or a much more seldom but very interesting effects
of pure state formation from mixed state input ---
spontaneous coherence,
exhibited in entropy decrease, synchronizing,
coherent responses of $\left(\psi_A+\psi_B\right)^2\gg \psi^2_A+\psi^2_B$ type,
etc.

Consider the density matrix mapping of $l=1\dots M$ observations
\begin{align}
\rho^{(l)}& \to \varrho^{(l)}
  & \text{weight $\omega^{(l)}$}  \label{mlproblemVectorDensityMatrix}
\end{align}
The $\rho^{(l)}$ is a measured Hermitian density matrix $n\times n$ in the $\Ket{\psi}$
space,
while
the $\varrho^{(l)}$ is a measured Hermitian density matrix $D\times D$ in the $\Ket{\phi}$ space.
Both $\rho$ and $\varrho$ density operators
are Hermitian matrices with positive eigenvalues 
and having  unit
\href{https://en.wikipedia.org/wiki/Trace_(linear_algebra)}{trace}
(the sum of diagonal elements).
Previously considered data (\ref{mlproblemVector})
corresponded to $\rho^{(l)}=\Ket{\psi_l}\Bra{\psi_l}$ and
$\varrho^{(l)}=\Ket{\phi_l}\Bra{\phi_l}$.
The optimization problem becomes
\begin{align}
\mathcal{F}&
=\sum\limits_{l=1}^{M} \omega^{(l)}
F\left(\varrho^{(l)},\sigma^{(l)}\right)
\xrightarrow[B]{\quad }\max
\label{FQuantumChannelMapping}
\end{align}
where $\sigma=A^{OUT}$ and $\rho=A^{IN}$ --- 
the $\sigma$ is the $\rho$ passed through the quantum channel.

It should be noted that while for wavefunction mapping (\ref{mlproblemVector}),
in the seldom case of a classic system with known wavefunctions phases,
a regression type technique is sometimes applicable,
for density matrix mapping (\ref{mlproblemVectorDensityMatrix}),
which is always quadratic on $\mathcal{U}$ (\ref{operatorTransform}),
no regression can possibly be applied.

We need to define 
the total fidelity $\mathcal{F}$ of the (\ref{KrausOperator}) transform,
which is a sum of contributions from all $M$ observations.
For each observation of the \emph{IN} to \emph{OUT}
mapping,
an accuracy factor $F$ ranging $[0:1]$ should be defined.
For the theory presented in this paper to be applicable,
this factor must be quadratic with respect to the operator $\mathcal{U}$ (an instance of $B_s$).
For pure state to pure state mapping (\ref{mlproblemVector})
it is a squared projection, the
\href{https://en.wikipedia.org/wiki/Fidelity_of_quantum_states}{fidelity}
\begin{align}
F&=\Big|\Braket{\phi|\mathcal{U}|\psi}\Big|^2
\label{QpureState}
\end{align}
For mixed state mapping (\ref{mlproblemVectorDensityMatrix})
a similar expression
\begin{align}
F^{\varrho\sigma}(\varrho,\sigma)&=\mathrm{Tr}\,
\varrho \sigma
\label{QpureStateDensityMatrix}
\end{align}
is bounded between $0$ and  $\mathrm{Tr}\varrho^2\le 1$,
it has the meaning of a measured probability of probability
that reaches the maximal value $1$ only in pure states.
A standard definition of fidelity between mixed states  $\varrho,\sigma$,
\cite{nielsen2010quantum} p. 409,
\cite{wilde2011classical} p. 285,
\begin{align}
F^{prop}(\varrho,\sigma)&=\mathrm{Tr}\sqrt{\varrho^{1/2}\sigma\varrho^{1/2}}=
\mathrm{Tr}\sqrt{\varrho}\sqrt{\sigma}
\label{fidelityStdDefinition}
\end{align}
creates calculation  difficulties since it requires taking matrix square root
of an expression with $B_s$ sum, and it is not explicitly quadratic.
There is an option to use a quality criterion based on the conditional entropy of $OUT|IN$ states \cite{jenvcova2024recoverability}.
In the general case, this requires a joint density matrix $\rho(OUT,IN)$ \cite{levitin1999conditional}.
The tensor  $S_{jk;j^{\prime}k^{\prime}}$ (\ref{StendorDensityMatrix})
can be viewed as a joint density matrix $\rho(OUT,IN)$ averaged over $M$ observations in the sample.
However, in the case where $N_s>1$, we were unable to express the conditional entropy in quadratic form with respect to $B_s$,
and we will leave this aspect of conditional entropy as a quality criterion for future research.
In the $N_s=1$ case, the Kullback-Leibler divergence (Eq. \ref{mutualEntropy}) can be directly applied.

There is a noticeable feature of Hermitian operators:
if one vectorizes $D\times D$ Hermitian operators
$\varrho_{ij}$ and $\sigma_{ij}$, then one can verify that
\begin{align}
\mathrm{Tr} \varrho \sigma &=
\sum\limits_{i,j=0}^{D-1} \varrho_{ij}\sigma^*_{ij}
\label{scalProdDM}
\end{align}
i.e. the $L^2$ of a vector obtained from all $D^2$ matrix elements of $\varrho$
gives the trace of $\varrho^2$.
This provides a closeness criterion, $F^{c}$, which acts as a ``correlation''
between the density matrices $\varrho_{ij}$ and $\sigma_{ij}$, treated as if they were vectors.
\begin{align}
F^{c}(\varrho,\sigma)&=\frac{\sum\limits_{i,j=0}^{D-1} \varrho_{ij}\sigma^*_{ij}}
{
\sqrt{
\sum\limits_{i,j=0}^{D-1} \left|\varrho_{ij}\right|^2
\sum\limits_{i,j=0}^{D-1} \left|\sigma_{ij}\right|^2
}
}=
\frac{\mathrm{Tr} \varrho \sigma}{
\sqrt{
\mathrm{Tr} \varrho^2
\,
\mathrm{Tr} \sigma^2
}
}
\label{FasCorrelation}
\end{align}
It can be viewed as $F^{\varrho\sigma}$ (\ref{QpureStateDensityMatrix})
with an adjusted contribution for mixed states.
The practical issue with it is that $\sigma$, in the denominator, depends on the quantum channel operators $B_s$;
to obtain a workable expression,
one can replace the $\sigma$-normalized denominator
by the input density matrix $\rho$
\begin{align}
F^{v}&=\frac{\sum\limits_{i,j=0}^{D-1} \varrho_{ij}\sigma^*_{ij}(\rho)}
{
\sqrt{\sum\limits_{i,j=0}^{D-1} \left|\varrho_{ij}\right|^2}
\sqrt{\sum\limits_{k,q=0}^{n-1} \left|\rho_{kq}\right|^2}
}
\label{fidelityScalProduct}
\end{align}
Here, the entity denoted as $\sigma(\rho)$ is obtained by passing $\rho$ through the quantum channel
using either the transformation in (\ref{KrausOperator}) or (\ref{operatorTransform}).
Since $\sigma$  does not appear in the denominator, Eq. (\ref{fidelityScalProduct})
is quadratic in the quantum channel mapping operators.
This expression uses (\ref{scalProdDM}) and treats density matrices as if they were vectors.
It is linear in $\sigma$  and quadratic in $\mathcal{U}$;
for unitary mapping ($D=n$, $N_s=1$), we have $\mathrm{Tr} \rho^2 = \mathrm{Tr} \sigma^2$,
and thus $F^v$ matches $F^{c}$.
An alternative is to consider a surrogate closeness by normalizing it only on $\varrho$ ---
to the degree of output state purity
\begin{align}
F^{N \varrho ^2}(\varrho,\sigma)&=\frac{\mathrm{Tr}\varrho \sigma}{\mathrm{Tr}\varrho^2} 
\label{QpureStateDensityMatrixNormalized}
\end{align}
but this one performs poorly, especially
in the case when $N_s>1$.
Note that if $\varrho$ is a pure state $\Ket{\phi}\Bra{\phi}$ 
and $\rho$ is a mixed state
then
(\ref{QpureStateDensityMatrix}),
(\ref{fidelityStdDefinition}),
and 
(\ref{QpureStateDensityMatrixNormalized}) are the same.
Whereas the closeness criteria
$F^{v}$ (\ref{fidelityScalProduct}) and $F^{N \varrho ^2}$ (\ref{QpureStateDensityMatrixNormalized})
are quadratic in quantum channel operators,
their main drawback is the lack of clear physical meaning.
This can lead to artifacts in the obtained solution,
particularly for noisy data in the case of a density matrix passing through a general quantum channel.
The criterion $F^{\varrho\sigma}$  (\ref{QpureStateDensityMatrix})
is quadratic in quantum channel operators but represents the square of a probability,
which overestimates the contribution of pure states\cite{malyshkin2019radonnikodym}.
As a result, it is poorly suited for density matrix mappings.
The criteria $F^{prop}$ (\ref{fidelityStdDefinition}) and $F^{c}$ (\ref{FasCorrelation})
have a clear physical meaning, but they are not quadratic in quantum channel operators.
Of particular interest is the criterion
$F^{\sqrt{\varrho}\sqrt{\rho}}$ (\ref{FSQRT}),
considered below,
which is quadratic in quantum channel operators and matches $F^{prop}$
exactly in the case of unitary mapping of density matrices.

For the theory presented in this paper to be applicable, the only required feature of the mapping accuracy
$F$ in (\ref{FQuantumChannelMapping})
is that it must be a \textsl{quadratic} form on
$B_s$
with a superoperator $S_{jk;j^{\prime}k^{\prime}}$.
\begin{align}
\mathcal{F}&=
\sum\limits_{s=0}^{N_s-1}
 \Braket{B_s\big|S\big|B_s}
 \label{FasOperatorsProduct}
\end{align}
The tensor $S_{jk;j^{\prime}k^{\prime}}$
depends on the choice of  $F$
and observation weighs $\omega^{(l)}$.
They are problem specific and 
do not change our considerations.
Expanding the sums in (\ref{FasOperatorsProduct}) we obtain
(for a simple  choice of $F$ (\ref{QpureStateDensityMatrix}))
\begin{align}
\mathcal{F}&=
\sum\limits_{l=1}^{M}\omega^{(l)}
\sum\limits_{s=0}^{N_s-1}
\sum\limits_{i,j=0}^{D-1}
\sum\limits_{k,k^{\prime}=0}^{n-1}
\varrho^{(l)}_{ij}b_{s,ik} \rho^{(l)}_{kk^{\prime}} b^*_{s,jk^{\prime}}
\label{Fexpansion}
\end{align}
where the $b_{s,ik}$ are $s=0\dots N_s-1$ matrices of
dimension  $D\times n$ corresponding to partially Kraus operators $B_s$.
One can introduce a tensor
\begin{align}
S_{jk;j^{\prime}k^{\prime}}&=
\sum\limits_{l=1}^{M}\omega^{(l)}
\varrho^{(l)}_{jj^{\prime}}\rho^{(l)}_{kk^{\prime}}
\label{StendorDensityMatrix}
\end{align}
to obtain an expression similar to the one in \cite{belov2024partiallyPRE}
for $\mathbf{x}\to\mathbf{f}$ Hilbert spaces vector mapping.
With the density matrix
$\Ket{\mathbf{x}}\Bra{\mathbf{x}}\to\Ket{\mathbf{f}}\Bra{\mathbf{f}}$
(\ref{StendorDensityMatrix}) takes a familiar form
\begin{align}
S_{jk;j^{\prime}k^{\prime}}&=
\sum\limits_{l=1}^{M} \omega^{(l)}
f^{(l)}_j x^{(l)}_k
f^{(l)\,*}_{j^{\prime}} x^{(l)\,*}_{k^{\prime}}
\label{SfunctionalClassic}
\end{align}
Distinguishing features of the current work include:
\begin{itemize}
\item Using density matrix states mapping (\ref{mlproblemVectorDensityMatrix})
as input.
\item Go beyond unitary mapping (\ref{operatorTransform})
to consider quantum channel mapping (\ref{KrausOperator}).
\end{itemize}
We use them to take into account
two kinds of probabilities:
probabilistic mixture of states and states superposition.
For the latter, probability is the square of the sum of amplitudes,
while for the former, probability is the sum of squared amplitudes.
Both effects are present in real life data.
There are a number of studies suggesting a limitation
of unitary scalarization \cite{kendall2018multi,sener2018multi},
the other  reevaluate recent research \cite{kurin2022defense}.
It should be noted how researchers commonly use unitarity ---
they take a vector, normalize its $L^2$ norm,
and consider the squared Euclidean projections obtained
as if they were probabilities.
Existing approaches do not
distinguish
probabilistic mixture of states from their superposition.
This work is trying to overcome
this deficiency by using input data
in the form of density matrices (\ref{mlproblemVectorDensityMatrix})
and mapping them with a general quantum channel (\ref{KrausOperator}).
Consider a few demonstrations.

The partially unitary mapping
(\ref{operatorTransform}) (when $N_s=1$) is a
\href{https://en.wikipedia.org/wiki/Quantum_channel}{trace decreasing map}
quantum channel for $D<n$. A simple $N_s>1$ example: Let $D=1$ and we want to construct
a quantum channel
with operators $B_s$
calculating the trace of $A^{IN}$ with (\ref{KrausOperator}).
The matrices $b_{s,jk}$ are of $1\times n$ dimension for all $s$.
Any orthogonal basis $\Ket{x_s}$ in $\Ket{\psi}$
solves the problem:
\begin{align}
B_s&=\Ket{f_0}\Bra{x_s}
\label{traceCalculatingB}
\end{align}
 with $N_s=n$,
$s=0\dots n-1$.
Any orthogonal basis $\Ket{x_s}$ 
creates a solution of the form (\ref{traceCalculatingB})
that satisfies the canonical form constraints (\ref{KrausCanonic}),
this degeneracy may cause difficulties in numerical methods.
Another example. A quantum channel parametrized by two orthogonal bases
$\Ket{x_k}$,
$\Ket{f_j}$, and a matrix $\mathcal{M}_{jk}$
creating $N_s=Dn$ rank one operators
\begin{align}
B_s&=\Ket{f_j}\mathcal{M}_{jk}\Bra{x_k}
\label{tracePreservingAnyDBM}
\end{align}
The index $s=0\dots Dn-1$ enumerates all $(j,k)$ pairs.
The actual Kraus rank is less than or equal to $Dn$.
The trace-preservation condition (\ref{constraintKraussSpur}) imposes $k=0\dots n-1$ constraints:
\begin{align}
1&=\sum\limits_{j=0}^{D-1}\left|\mathcal{M}_{jk}\right|^2
\label{tracePreservingAnyDBMConstr}
\end{align}
The matrix $\mathcal{M}_{jk}$ defines a quantum channel.
Equations (\ref{traceCalculatingB}) and (\ref{tracePreservingAnyDBM})
are two examples of trace-preserving quantum channels,
constructed as a sum of rank one operators
$\Ket{f_{j}}\Bra{x_k}$.
Such channels are defined using two bases, $\Ket{f_{j}}$, $\Ket{x_k}$,
along with the mapping matrix $\mathcal{M}_{jk}$ between them.
These channels are much easier to analyze (and work numerically)
compared to a sum of general operators $B_s$.
At the same time, they possess sufficiently high expressive
power for trace-type mappings.
However, a unitary mapping cannot be represented by (\ref{tracePreservingAnyDBM}).

For a general form of quantum channel see \cite{torlai2023quantum,duvenhage2021optimal}
for a representation of a quantum channel with Choi matrix of the channel\cite{choi1975completely}.
Let us defer the study of this Choi-style
$J(\Phi)=\sum_{kk^{\prime}} \Ket{k}\Bra{k^{\prime}} \otimes \Phi(\Ket{k}\Bra{k^{\prime}})$
representation
of a
\href{https://www.youtube.com/watch?v=cMl-xIDSmXI}{quantum channel} $\Phi$
to future research and focus on the original problem.

Mathematically, the problem becomes: optimize (\ref{KrausMax})
subject to (\ref{KrausOrt1}) and (\ref{KrausCanonic}) constraints.
\begin{align}
\mathcal{F}&=
\sum\limits_{s=0}^{N_s-1}
\sum\limits_{j,j^{\prime}=0}^{D-1}\sum\limits_{k,k^{\prime}=0}^{n-1}
b_{s,jk}S_{jk;j^{\prime}k^{\prime}}b^*_{s,j^{\prime}k^{\prime}}
\xrightarrow[{b}]{\quad }\max
\label{KrausMax} \\
\delta_{jj^{\prime}}&=
\sum\limits_{s=0}^{N_s-1}
\sum\limits_{k=0}^{n-1}b_{s,jk}b^*_{s,j^{\prime}k}
\text{\qquad\quad $j,j^{\prime}=0\dots D-1$}
    \label{KrausOrt1}
\end{align}
A selection of the gauge is required to avoid problem degeneracy,
for example take the canonical form.
\begin{align}
0&=    
\sum\limits_{j=0}^{D-1}\sum\limits_{k=0}^{n-1} b_{s,jk} b^*_{s^{\prime},jk}
& s\ne s^{\prime} ,s=0\dots N_s-1
\label{KrausCanonic}
\end{align}
Other gauges
such as
\href{https://en.wikipedia.org/wiki/Cholesky_decomposition}{Cholesky decomposition} (on index $s$) can possibly be used instead.

The main feature of the numerical method we developed
consists of
using eigenvalue problem as the algorithm's building block.
If we consider a subset of (\ref{KrausOrt1}) constraints,
then the optimization problem can be readily solved.
Consider
the squared
\href{https://en.wikipedia.org/wiki/Matrix_norm#Frobenius_norm}{Frobenius norm}
of $b_{s,jk}$
to be a ``simplified constraint'':
\begin{align}
\sum\limits_{s=0}^{N_s-1}
\sum\limits_{j=0}^{D-1}
\sum\limits_{k=0}^{n-1}\left|b_{s,jk}\right|^2
&=D
    \label{KrausPartialConstraint}
\end{align}
This is a partial constraint
(it is the sum of all diagonal elements in (\ref{KrausOrt1})).
For this partial constraint the optimization problem (\ref{KrausMax})
is equivalent to an eigenvalue problem  --- it can be directly solved
by considering a vector of dimension $N_sDn$ 
obtained from the $b_{s,jk}$ operator by 
saving all its components to a
\href{https://en.wikipedia.org/wiki/Vectorization_(mathematics)}{single vector},
row by row.
With $s$-independent $S_{jk;j^{\prime}k^{\prime}}$  each eigenvalue is degenerate $N_s$ times;
the Lagrange multipliers $\nu_{ss^{\prime}}$ in (\ref{KrausSwithLagrange}) may potentially remove this degeneracy.
This partially constrained problem
is the main building block of our numerical algorithm.
Whereas most existing learning algorithms use either first order gradient-style methods or second order Newtonian methods for optimization,
an eigenvalue problem is the building block of the algorithm in this paper.\footnote{
We think that the reasons why first order gradient methods are preferred in neural networks over second order Newtonian-type methods are:
1. The problem may be of very high dimension,
and first order methods do not need to store a large Hessian matrix.
2. The biggest problem in learning optimization is not local maximums, but saddle points\cite{arjovsky2015saddle}.
First order methods are less likely to get stuck in a saddle point.
3. For some tasks, first-order gradient-type methods may offer better computational complexity,
see Appendix \ref{compComplexityEstimation} below.
}
This represents a transition from mathematical analysis tools (e.g., gradient, derivative, etc.) to using algebraic tools (eigenproblem).
This transition enables the move from single solution methods
to multiple solutions (eigenvectors).
This transition from analysis to algebra
makes finding the global maximum
much more likely, not to mention providing a better understanding
of the problem itself.

\section{\label{UnitaryMappingOfMixedStates}A Unitary Mapping of Mixed States}
In quantum channel learning there are two sources of mixed states.
First, the original input density matrices (\ref{mlproblemVectorDensityMatrix})
can be in a mixed state.
Secondly, the quantum channel itself, through the transformation
 (\ref{KrausOperator}),
can create mixed states.
In this section we consider the first source.
Let the data be of mixed states according to mapping
(\ref{mlproblemVectorDensityMatrix}),
while the quantum channel itself is assumed to be unitary
(\ref{operatorTransform})
or possibly partially unitary with $D\le n$.
To apply our theory and software we need to obtain
a tensor $S_{jk;j^{\prime}k^{\prime}}$ with which to solve the optimization problem
(\ref{KrausMax}), now with $N_s=1$.
\begin{align}
\mathcal{F}&=
\sum\limits_{j,j^{\prime}=0}^{D-1}\sum\limits_{k,k^{\prime}=0}^{n-1}
u_{jk}S_{jk;j^{\prime}k^{\prime}}u^*_{j^{\prime}k^{\prime}}
\xrightarrow[{u}]{\quad }\max
\label{FUMax} \\
\delta_{jj^{\prime}}&= \sum\limits_{k=0}^{n-1}u_{jk}u^*_{j^{\prime}k}
\text{\qquad\quad $j,j^{\prime}=0\dots D-1$}
    \label{FUMaxConstraints}
\end{align}
If we take the tensor (\ref{StendorDensityMatrix}) as $S_{jk;j^{\prime}k^{\prime}}$ ---
then we underestimated fidelity, 
which leads to unusual effects,
see Table \ref{tabHierarachy} row $F^{\varrho\sigma}$.

To obtain a proper estimation of the closeness of states for density matrix input, consider the mapping of the square roots of density matrices.
\begin{align}
\sqrt{\rho^{(l)}}& \to \sqrt{\varrho^{(l)}}
  & \text{weight $\omega^{(l)}$}  \label{mlproblemVectorDensityMatrixSQRT}
\end{align}
and use it instead of (\ref{mlproblemVectorDensityMatrix}),
as if it were the actual density matrix mapping.
For a unitary mapping $\mathcal{U}$
the same quantum channel converts both the density matrix and its square root.
If $\sqrt{\varrho}=\mathcal{U} \sqrt{\rho} \mathcal{U}^{\dagger}$
then $\varrho=\mathcal{U} \sqrt{\rho} \mathcal{U}^{\dagger} \mathcal{U} \sqrt{\rho} \mathcal{U}^{\dagger}=
\mathcal{U} \rho \mathcal{U}^{\dagger}$;
also see Exercise 9.14 from \cite{nielsen2010quantum}, p. 410:
for any positive operator $A$, $\sqrt{UAU^{\dagger}}= U\sqrt{A}U^{\dagger}$.
Thus we can use\footnote{
\label{infiltrationProp}
Note that from the invariance property of unitaries,
$Uf(A)U^{\dagger}=f(UAU^{\dagger})$,
for any positive operator $A$ and $p\ge 0$, we have $\left(UAU^{\dagger}\right)^p= UA^pU^{\dagger}$.
Thus, a form of
$S_{jk;j^{\prime}k^{\prime}}=
\sum_{l=1}^{M}\omega^{(l)}
\left(\varrho^{(l)}\right)^p_{jj^{\prime}}\left(\rho^{(l)}\right)^q_{kk^{\prime}}
$
with $p+q=1$
can be considered.
However, an arbitrary $p$ and $q$ does not correspond to the fidelity in Eq. (\ref{fidelityStdDefinition}), which has $p=q=1/2$,
resulting in Eq. (\ref{StendorDensityMatrixSQRTrho}).
For an arbitrary $N_s$, any form with $q=1$ can be reduced to a QCQP optimization problem.
Equation (\ref{QpureStateDensityMatrix}) corresponds to $p=q=1$.
One can consider a form with $q=1$ and some $p$  as a quadratic form proxy for the general quantum channel fidelity.
}
\begin{align}
S_{jk;j^{\prime}k^{\prime}}&=
\sum\limits_{l=1}^{M}\omega^{(l)}
\left(\sqrt{\varrho^{(l)}}\right)_{jj^{\prime}}\left(\sqrt{\rho^{(l)}}\right)_{kk^{\prime}}
\label{StendorDensityMatrixSQRTrho}
\end{align}
in the optimization problem
if we apply the closeness (\ref{QpureStateDensityMatrix})
to the mapping of density matrix square root
(\ref{mlproblemVectorDensityMatrixSQRT}).
Let us denote it as $F^{\sqrt{\varrho}\sqrt{\rho}}$.
\begin{align}
F^{\sqrt{\varrho}\sqrt{\rho}}&=\mathrm{Tr}\sqrt{\varrho}\sigma(\sqrt{\rho})
\label{FSQRT}
\end{align}
Here the entity denoted as $\sigma(\sqrt{\rho})$ 
is obtained by passing $\sqrt{\rho}$
through the quantum channel using either the transformation in (\ref{KrausOperator}) or (\ref{operatorTransform}).
In the case (\ref{operatorTransform}), where $N_s=1$, this $F^{\sqrt{\varrho}\sqrt{\rho}}$
is equal to the proper fidelity $F^{prop}$ (\ref{fidelityStdDefinition}).
The approach creates no difficulty in implementation,
as the square roots of density matrices are calculated upfront and then
used as if they were actual measurements.
The $S_{jk;j^{\prime}k^{\prime}}$  from (\ref{StendorDensityMatrixSQRTrho})
corresponds to the density matrix square root mapping in (\ref{mlproblemVectorDensityMatrixSQRT}).
For a pure state $\rho=\Ket{\psi}\Bra{\psi}$, we have $\rho=\sqrt{\rho}$,
and (\ref{FSQRT}) gives the same result as (\ref{QpureState}).

\subsection{\label{infbased}A Conditional Entropy Based Similarity}
In the case of density matrix unitary mapping, there is a possible alternative to the square root mapping
$\sqrt{\rho} \to \sqrt{\varrho}$ (\ref{mlproblemVectorDensityMatrixSQRT})
that has been considered above.
Now consider the mutual information between the density matrices $\varrho$ and $\sigma$: the
\href{https://en.wikipedia.org/wiki/Kullback%E2%80%93Leibler_divergence}{Kullback-Leibler divergence}\cite{van2014renyi,galke2024sufficiency,lipka2024catalysis}.
Given two density matrices $\varrho$ and $\sigma$, the conditional entropy is
\begin{align}
S(\varrho||\sigma)&=
\mathrm{Tr}( \varrho \ln \varrho - \varrho \ln \sigma)
\label{mutualEntropy}
\end{align}
An interpretation of the Kullback Leibler divergence of $\varrho$ from $\sigma$
is the expected excess surprise from using $\sigma$ as a model instead of $\varrho$ when the actual distribution is $\varrho$.
It satisfies the Gibbs inequality:
$S(\varrho||\sigma) \ge 0$, $S(\varrho||\sigma)=0$ iff  $\varrho=\sigma$.
For $\sigma$ obtained as a unitary mapping (\ref{operatorTransform}) from $\rho$,
the $\ln \sigma = \ln  \left( U \rho U^{\dagger}\right)$
can be calculated using the \hyperref[infiltrationProp]{invariance property} of the unitaries.
Taking into account that $\mathrm{Tr}( \varrho \ln \varrho)$ is a constant one may consider the mapping
\begin{align}
\ln \rho^{(l)}& \to \varrho^{(l)}
  & \text{weight $\omega^{(l)}$}  \label{mlproblemVectorDensityMatrixLogRho}
\end{align}
and use the tensor
\begin{align}
S_{jk;j^{\prime}k^{\prime}}&=
\sum\limits_{l=1}^{M}\omega^{(l)}
\left(\varrho^{(l)}\right)_{jj^{\prime}}\left(\ln \rho^{(l)}\right)_{kk^{\prime}}
\label{StendorDensityMatrixLogRho}
\end{align}
as if it were the actual density matrix mapping.
With this $S_{jk;j^{\prime}k^{\prime}}$, the functional $\mathcal{F}$ no longer
has the meaning of the number of observations,
and special care should be taken when regularizing $\rho$ states with eigenvalues equal to zero due to logarithm calculations,
the $\varrho$ must also be zero in this state.
This information divergence measure is expected to be a good alternative
to the square root mapping (\ref{StendorDensityMatrixSQRTrho}),
but it requires more research, especially regarding the possibility of generalizing it to a quadratic form
for quantum channels with Kraus rank $N_s>1$
and its property of not satisfying the triangle inequality.

\subsection{\label{sqrtMappingDemo}A Demonstration Of Density Matrix Square Root Mapping}
Let us demonstrate the advantages of using the square root of the density matrix in quantum channel learning.
Here and below we need a number of random
density matrix input states $\rho^{(l)}$
(of a given rank  $N_r$, all real for simplicity)
that we map with the quantum channel of form
(\ref{operatorTransform}) or (\ref{KrausOperator}) to $\varrho^{(l)}$.
The states are created from random vectors $v^{(l)}_{r,k}$ as follows:
\begin{align}
\rho^{(l)}_{kk^{\prime}}&=\frac{1}{\mathrm{Norm}^{(l)}}\sum\limits_{r=0}^{N_r-1} v^{(l)}_{r,k} v^{(l)}_{r,k^{\prime}}
\label{randomRho}
\end{align}
For every $l$ we generate $N_r$ random real vectors $v^{(l)}_{r,k}$ of dimension $n$.
The density matrix
is obtained as a $r=0\dots N_r-1$ sum of
the \href{https://en.wikipedia.org/wiki/Dyadics#Dyadic,_outer,_and_tensor_products}{dyadic product} of vector $v^{(l)}_{r,k}$ with itself,
and then it is normalized to $1=\sum_{k=0}^{n-1}\rho^{(l)}_{kk}$.
These $\rho^{(l)}_{kk^{\prime}}$ are used as input density matrices
in our numerical experiments.

\begin{figure}
\includegraphics[width=0.7\columnwidth]{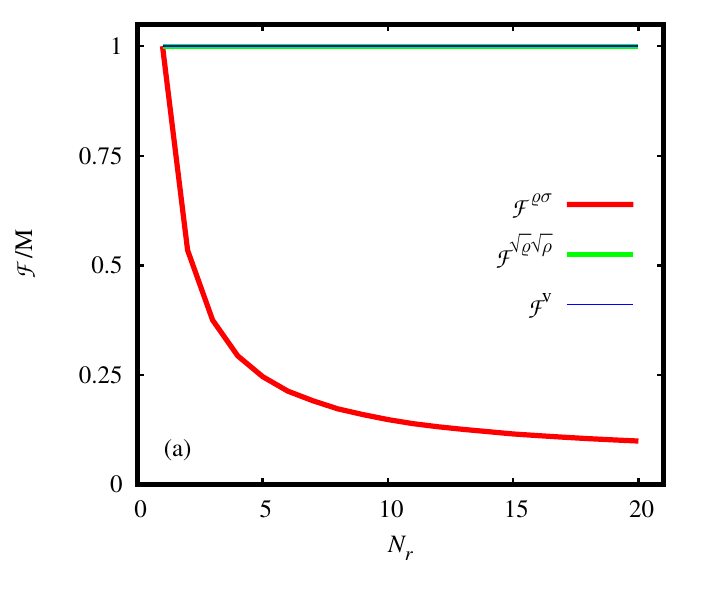}
\includegraphics[width=0.7\columnwidth]{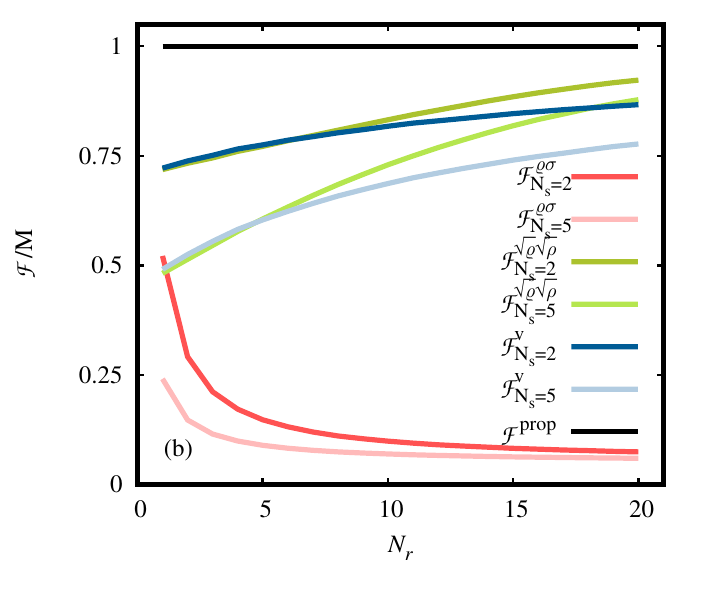}
%FMMEM SET width=0.65 for single column layout and width=0.95 for double column
\caption{\label{SkSQRTFig}
\baselineskip16pt
For known density matrix mappings ($\mathcal{F}^{prop}=M=1000$)
evaluate the total fidelity $\mathcal{F}$ (\ref{FQuantumChannelMapping})
(with different definitions of closeness $F$)
as a function
of the rank $N_r$
of the input density matrix
and the quantum channel's Kraus rank $N_s$.
Problem dimension is $n=D=20$.
(a) Unitary mapping $N_s=1$:
The
 $\mathcal{F}^{\varrho\sigma}$ (\ref{QpureStateDensityMatrix})
strongly depends on the rank of the input density matrices.
The $\mathcal{F}^{\sqrt{\varrho}\sqrt{\rho}}$ (corresponding to (\ref{mlproblemVectorDensityMatrixSQRT})
mapping)
and $\mathcal{F}^v$ (\ref{fidelityScalProduct})
produce the exact result $\mathcal{F}=M$
in the unitary mapping case.
(b) Multiple terms $N_s>1$:
The $\mathcal{F}^{\varrho\sigma}$ decreases even more strongly,
while
$\mathcal{F}^{\sqrt{\varrho}\sqrt{\rho}}$ and $\mathcal{F}^v$ are no longer exact.
}
\end{figure}

A demonstration is performed as follows. We generate
a quantum channel (\ref{KrausOperator}) of dimension $n=D=20$
with a given Kraus rank $N_s$.
Then for $N_r=1\dots 20$,
we generate $M=1000$ input
density matrices $\rho^{(l)}_{kk^{\prime}}$ (\ref{randomRho})
of rank $N_r$
and for each of them, we obtain the output
density matrix $\varrho^{(l)}_{jj^{\prime}}$ by passing
$\rho^{(l)}_{kk^{\prime}}$ through the quantum channel.
On this $\rho^{(l)}_{kk^{\prime}}\to\varrho^{(l)}_{jj^{\prime}}$
mapping we evaluate the total fidelity $\mathcal{F}$ (\ref{FQuantumChannelMapping})
for different definitions of $F$
(with the same $B_s$ defining the quantum channel,
$\sigma=\sum_{s=0}^{N_s-1} B_s \rho B^{\dagger}_s$).
The result is presented in Fig. \ref{SkSQRTFig};
see \texttt{\seqsplit{com/polytechnik/algorithms/DemoDMsqrtRhoMappingTest.java}} for an implementation.
For the $F^{prop}$ (\ref{fidelityStdDefinition})
the total fidelity (\ref{FQuantumChannelMapping}) is equal to the maximal value $\mathcal{F}^{prop}=M$
since the same $B_s$ were used both in construction and in evaluation
of the quantum channel.
However, a QCQP optimization problem that we can solve,
an algebraic problem (\ref{eigenvaluesLikeProblem}),
must be a quadratic function on $\mathcal{U}$.
The $F^{prop}$ is not such a function,
which makes it problematic to use.
A simple $F^{\varrho\sigma}$ as density matrices trace (\ref{QpureStateDensityMatrix})
is a quadratic function, but it is not normalized
for $N_r>1$ or $N_s>1$:
it produces the exact result only for 
pure state input ($N_r=1$) and unitary ($N_s=1$)  mappings,
see Fig. \ref{SkSQRTFig}a.
The mapping $\sqrt{\rho^{(l)}_{kk^{\prime}}}\to\sqrt{\varrho^{(l)}_{jj^{\prime}}}$
allows us to introduce $F^{\sqrt{\varrho}\sqrt{\rho}}$ (\ref{FSQRT})
considering $\sqrt{\varrho}$ and $\sqrt{\rho}$ as if they were actually
measured density operators (\ref{mlproblemVectorDensityMatrixSQRT}).
In the unitary case $N_s=1$ it
produces the correct value $\mathcal{F}=M$ 
for any rank of the input density matrix $\rho$,
this is an important result of this work.
In this $N_s=1$ case
for each $N_r$ we also run the operator $\mathcal{U}$ reconstruction algorithm
of Appendix \ref{numericalSolutionSect}
--- it is correctly (with zero error)  recovered (up to a sign)
for any $N_r$ and any measure of closeness used:
$F^{\varrho\sigma}$,
$F^v$,
$F^{N \varrho ^2}$,
or
$F^{\sqrt{\varrho}\sqrt{\rho}}$, 
regardless of the specifics of $F$.
This occurs only for exact mapping;
when a noise is present in the data --- the low values
of $F^{\varrho\sigma}$ make it problematic to reconstruct
the quantum channel from input density matrices.
The $F^{\sqrt{\varrho}\sqrt{\rho}}$ does not have this limitation
as for $N_s=1$ it is equal to proper fidelity $F^{prop}$ (\ref{fidelityStdDefinition}).
We see $F^{\sqrt{\varrho}\sqrt{\rho}}$ as a very
promising approach to the problem
of reconstructing a unitary operator from a
set of mixed states density matrix mapping.
Another one if $F^{v}$ (\ref{fidelityScalProduct}), which treats density matrices
as ``vectors'', but it does not have a clear physical meaning.

The situation with quantum channels of Kraus rank
$N_s>1$ is more problematic.
The
$F^{\varrho\sigma}$
strongly underestimates the fidelity, and
the $F^{\sqrt{\varrho}\sqrt{\rho}}$ is no longer correct because when there is more than one term
in the sum of (\ref{KrausOperator}), no quantum channel simultaneously converts $\rho$ and $\sqrt{\rho}$ between Hilbert spaces.
The main problem we encountered is constructing a fidelity that can be expressed as a quadratic function of $B_s$ in the form of (\ref{FasOperatorsProduct}).
Not every fidelity can be expressed in this form.
Using a proxy, such as $F^{\varrho\sigma}$, $F^{N \varrho ^2}$, etc.,
creates difficulties -- for example
it may become possible to obtain a fidelity value
greater than the one calculated with the exact mapping of the quantum channel.
This is the main requirement  for a ``fidelity proxy'':
it should reach the maximal value at the exact mapping of the quantum channel,
see Section \ref{DemoHierarchy} below for counterexamples.

\section{\label{Kraushierarchy}A Hierarchy of Unitary Operators}
As discussed below in 
Appendix \ref{numericalSolutionSect},
a direct numerical attempt to obtain Kraus operators
(\ref{KrausOperator})
in canonical gauge (\ref{KrausCanonic})
does not work due to the degeneracy caused
by the increased eigenproblem dimension
from $Dn$ to $N_sDn$
and the difficulty in formulating convergence helper constraints.
We previously attempted to use a completely different approach for finding Kraus operators that involved a special
parametrization of operators.
However, it turned out to be not very stable numerically and applicable only to problems of small dimensions, $n\lesssim 3$,
such as
\href{https://en.wikipedia.org/wiki/3D_rotation_group}{SO(3)},
see \cite{schafers2024modified}  for a similar consideration.

Let us approach the problem of quantum channel reconstruction
from a ``numerical algorithm'' perspective.
What quantum channel problem we can  efficiently solve
for high dimension, i.e. $D\le n>50$?
Actually --- only the problem of finding the optimal
unitary mapping (\ref{operatorTransform}) that maximizes
 (\ref{FUMax})
subject to  partial unitarity constraints (\ref{FUMaxConstraints}).

Our algorithm
\cite{belov2024partiallyPRE}
has good stability and convergence. For an updated version
see Appendix \ref{numericalSolutionSect} and set $N_s=1$ in all formulas.
Now, instead of optimizing the Kraus problem  (\ref{KrausMax})
subject to constraints (\ref{KrausOrt1}) and (\ref{KrausCanonic}),
consider the unitary hierarchy.
A quantum channel is built as\footnote{
This quantum channel
has an interesting physical interpretation.
If (\ref{UnitaryHierarchyQC})
represents a time evolution, then one may think of it as a quantum system evolving with several Hamiltonians at once,
$\mathcal{U}^{[s]}=\exp \left[-i\frac{t}{\hbar} H_s \right]$,
rather than as a system evolving with the single Hamiltonian $H=\sum\limits_s H_s$.
}
\begin{align}
  A^{\emph{OUT}}&=\sum\limits_{s=0}^{N_s-1} |w_s|^2 \mathcal{U}^{[s]} A^{\emph{IN}} \mathcal{U}^{[s]\,\dagger}
  \label{UnitaryHierarchyQC} \\
  1&=\sum\limits_{s=0}^{N_s-1}|w_s|^2 \label{wNorm1}
\end{align}
where $\mathcal{U}^{[s]}$ are partially unitary operators
satisfying the constraints (\ref{FUMaxConstraints})
and $|w_s|^2$ are positive real weights.
Since the weights are positive,
the channel is a convex combination of unitary channels ---
a \href{https://learning.quantum.ibm.com/course/general-formulation-of-quantum-information/quantum-channels#convex-combinations-of-channels}{mixed unitary channel}.
Note that not every quantum channel can be represented as a mixed-unitary channel\cite{lee2020detecting}.
The authors \cite{girard2022mixed} proved that for every mixed-unitary channel,
the following rank inequality holds between the Kraus rank $N_s^K$ (\ref{KrausOperator})
and the mixed-unitary rank $N_s^U$ (the number of terms in (\ref{UnitaryHierarchyQC})).
\begin{align}
N_s^U&\le\left(N_s^K\right)^2 -N_s^K +1
\label{NKNU}
\end{align}
It may look as the sum (\ref{UnitaryHierarchyQC})
exactly represents the Kraus quantum channel 
(\ref{KrausOperator}) with  $B_s=w_s \mathcal{U}^{[s]}$
(they are not necessary in the canonical form (\ref{KrausCanonicForm}),
see Appendix \ref{KrausCanonicMethod} for a transformation),
but actual Kraus rank can be lower than the number of terms in the sum (\ref{UnitaryHierarchyQC}).
However,
if we replace the sum of general Kraus operators $B_s$ by
a sum of partially unitary operators $\mathcal{U}^{[s]}$,
then we can apply our numerical algorithm to build the quantum channel
as a hierarchy of partially unitary operators.
Considering this convex combination of unitary channels
we can solve the problem incrementally.
This is the cost required to apply our numerical method\cite{belov2024partiallyPRE}
to quantum channel reconstruction.

The idea of finding $\mathcal{U}^{[s]}$ is similar to
density matrix reconstruction. Assume we have a density matrix $\varrho$
of dimension $N_s$
with some eigenvectors  $\phi^{[s]}$ and eigenvalues $P^{[s]}$,
a convex combination of pure states.
\begin{align}
\varrho &= \sum\limits_{s=0}^{N_s-1} P^{[s]} \Ket{\phi^{[s]}} \Bra{\phi^{[s]}}
\label{densMatr} \\
\Ket{\varrho\middle|\phi^{[s]}}&=P^{[s]} \Ket{\phi^{[s]}} \label{densMatrEigenproblem}
\end{align}
To recover
 $\Ket{\phi^{[s]}}$ and $P^{[s]}$
one may solve a sequence of $N_s$
constrained optimization problems.
\begin{align}
P&=\frac{
\Braket{\phi|\varrho|\phi}
}{
\Braket{\phi|\phi}
}\xrightarrow[{\mathcal{\phi}}]{\quad }\max \label{DensMatrixMax} \\
0&=\Braket{\phi|\phi^{[s^{\prime}]}} & s^{\prime}<s \label{constraintsDensMatrix}
\end{align}
On each step a pair $\left(P^{[s]},\Ket{\phi^{[s]}}\right)$
is obtained
from the optimization problem (\ref{DensMatrixMax})
subject to the homogeneous linear constraints (\ref{constraintsDensMatrix}).
The probabilities $P^{[s]}$ decrease with $s$ and
the obtained pairs form a hierarchy of states from most probable
to less probable.

In \cite{malyshkin2019radonnikodym,malyshkin2022machine,belov2024partiallyPRE}
we formulated a novel algebraic problem
\begin{align}
  S \mathcal{U} &= \lambda \mathcal{U}
  \label{eigenvaluesLikeProblem}
\end{align}
where superoperator $S$ is a tensor
$S_{jk;j^{\prime}k^{\prime}}$, 
``eigenvector'' $\mathcal{U}$ is a partially unitary
operator (\ref{operatorTransform})
represented by
a $D\times n$ matrix $u_{jk}$,
and ``eigenvalue''  $\lambda$ is a Hermitian $D\times D$ matrix
of Lagrange multipliers $\lambda_{ij}$ (\ref{newLambdaSolPartial}).
See Appendix \ref{SchodingerNonStationary} below for its time-dependent form (\ref{SchodingerNonStationaryEq}).

To construct a hierarchy
of (Lagrange multipliers, operator) pairs $\left(\lambda^{[s]},\mathcal{U}^{[s]}\right)$
of decreasing fidelity  $\mathcal{F}^{[s]}=\mathrm{Tr} \lambda^{[s]}$
we need an analogue of the already found states' orthogonality
condition.
Whereas in  (\ref{constraintsDensMatrix})
for a  regular eigenproblem with a scalar eigenvalue
the result is the same regardless of performing an inner product
with the ``numerator'' $P^{[s]}\Ket{\phi^{[s]}}$
or ``denominator'' $\Ket{\phi^{[s]}}$ terms,
in the algebraic problem (\ref{eigenvaluesLikeProblem}) with $\lambda_{ij}$
being Hermitian matrix this is no longer the case.
We have a number of options for orthogonality conditions
for already found states $s^{\prime}<s$,
for example (\ref{constraintsUPreviousNumerator}) is an analogue of the ``numerator''
and (\ref{constraintsUPreviousDeniminator}) of the ``denominator''.
\begin{subequations}
\begin{align}
0&=\sum\limits_{i,j=0}^{D-1}\sum\limits_{k=0}^{n-1} u_{ik} \lambda^{[s^{\prime}]}_{ij} u^{[s^{\prime}]\,*}_{jk}
=\Braket{\mathcal{U}|\lambda^{[s^{\prime}]}|\mathcal{U}^{[s^{\prime}]}}
\label{constraintsUPreviousNumerator} \\
0&=\sum\limits_{j=0}^{D-1}\sum\limits_{k=0}^{n-1} u_{jk} u^{[s^{\prime}]\,*}_{jk}
=\Braket{\mathcal{U}|\mathcal{U}^{[s^{\prime}]}}
\label{constraintsUPreviousDeniminator}
\end{align}
\end{subequations}
However the most convenient constraints correspond
to a hierarchy of orthogonal quantum channels:
\begin{align}
0&=\sum\limits_{j,j^{\prime}=0}^{D-1}\sum\limits_{k=k^{\prime}=0}^{n-1}
u_{jk}
S_{jk;j^{\prime}k^{\prime}}
u^{[s^{\prime}]\,*}_{j^{\prime}k^{\prime}}
=\Braket{\mathcal{U}|S|\mathcal{U}^{[s^{\prime}]}}
\label{constraintsUPreviousOrthogonalChannel}
\end{align}
Our numerical algorithm allows any homogeneous linear constraint to be incorporated into (\ref{KrauslinearConstraintsHomog}).
We believe that the appropriate orthogonality constraint on the previous state is the homogeneous linear constraint (\ref{constraintsUPreviousOrthogonalChannel}).
We also use the ``denominator''-type constraint (\ref{constraintsUPreviousDeniminator})
in Appendix \ref{SchodingerNonStationary} below to construct a density tensor.
The reason why we choose the (\ref{constraintsUPreviousOrthogonalChannel}) form ---
assume an operator $\mathcal{V}$ is built as
\begin{align}
\mathcal{V}&=\sum\limits_{s=0}^{N_s-1} w_s \mathcal{U}^{[s]}
\label{VSumUs}
\end{align}
Then, using the constraint (\ref{constraintsUPreviousOrthogonalChannel}),
obtain
\begin{align}
\Braket{\mathcal{V}|S|\mathcal{V}}&=
\sum\limits_{s=0}^{N_s-1} |w_s|^2 \Braket{\mathcal{U}^{[s]}|S|\mathcal{U}^{[s]}}
\label{qcAsSum}
\end{align}
i.e. a single nonunitary operator $\mathcal{V}$
can completely define a quantum channel (\ref{UnitaryHierarchyQC}) with $N_s>1$.
For a given $\mathcal{V}$ the expansion weights in (\ref{VSumUs})
are obtained as
\begin{align}
w_s&=
\frac{\Braket{\mathcal{V}|S|\mathcal{U}^{[s]}}}
{\Braket{\mathcal{U}^{[s]}|S|\mathcal{U}^{[s]}}}
\label{wsExpr}
\end{align}
A hierarchy of (Lagrange multipliers, operator) pairs
is constructed by performing (\ref{FUMax})
partially unitary optimization (not Kraus!)
$N_s$ times.
For each $s$ include the constraints (\ref{constraintsUPreviousOrthogonalChannel})
from all previous  optimizations $s^{\prime}=0\dots s-1$ 
into the full set (\ref{KrauslinearConstraintsHomog}).
The result of this procedure is a hierarchy of $N_s$
pairs $\left(\lambda^{[s]}_{ij},u^{[s]}_{jk}\right)$
with decreasing fidelity.

The tensor 
$S_{jk;j^{\prime}k^{\prime}}$
can be expanded:
\begin{align}
S&\approx\sum\limits_{s=0}^{N_s-1} \frac{1}{\mathcal{F}^{[s]}}
\Ket{S\middle|\mathcal{U}^{[s]}}
\Bra{\mathcal{U}^{[s]}\middle| S }
\label{expansionSinUs}\\
 \mathcal{F}^{[s]}&=\sum_{i=0}^{D-1} \lambda_{ii}^{[s]}=
 \Braket{\mathcal{U}^{[s]}\big|S\big|\mathcal{U}^{[s]}}
 \label{FsHigh} \\
  D&=\Braket{\mathcal{U}^{[s]}\big|\mathcal{U}^{[s]}} \label{normD} \\
 \mathcal{F}^{[s]}\delta_{ss^{\prime}}&= \Braket{\mathcal{U}^{[s]}\big|S\big|\mathcal{U}^{[s^{\prime}]}}
 \label{FexpansionOrt}
\end{align}
This  expansion
is similar to the eigenvector expansion (\ref{densMatr}),
but ket $\Ket{\cdot}$ and bra $\Bra{\cdot}$ are now operators,
and $S$ is called a
\href{https://en.wikipedia.org/wiki/Superoperator}{superoperator}.
Note that usual orthogonality does not hold
$\Braket{\mathcal{U}^{[s]}\big|\mathcal{U}^{[s^{\prime}]}}\ne\delta_{ss^{\prime}}$,
(\ref{normD}) follows from the partial unitarity constraints (\ref{FUMaxConstraints}),
and (\ref{FexpansionOrt}) follows from the hierarchy constraints (\ref{constraintsUPreviousOrthogonalChannel}).
By continuously solving ($s=0,1,2,\dots$) the optimization problem (\ref{FUMax})
subject to the constraints
(\ref{FUMaxConstraints}) and (\ref{constraintsUPreviousOrthogonalChannel})
we can build the required hierarchy
of solutions
$\left(\lambda^{[s]}_{ij},u^{[s]}_{jk}\right)$
to reconstruct the quantum channel from it,
see 
\texttt{\seqsplit{com/polytechnik/kgo/KGOHierarchy.java}}
for an implementation.
A possible issue with the constraint (\ref{constraintsUPreviousOrthogonalChannel})
is that for the solution $s$ in the hierarchy,
Eq. (\ref{eigenvaluesLikeProblem}) is not satisfied for some projections,
the number of which equals the number of previous states in the hierarchy;
for the ground state $s=0$, all projections are satisfied.
However, the main advantage of (\ref{constraintsUPreviousOrthogonalChannel})
is that the matrix (\ref{FexpansionOrt})
for an approximated $S$ from (\ref{expansionSinUs}) with the full $N_s=Dn$ basis
is equal to the corresponding matrix with the exact $S$.
For the reconstruction of a quantum channel as a mixed unitary channel,
a few solutions with high $\mathcal{F}^{[s]}$ may be sufficient in a number of practical ML problems.
The obtained hierarchy may have up to $Dn$ solutions,
which is the maximum possible number $N_s$ of terms in the Kraus sum (\ref{KrausOperator}).

An application of the constructed hierarchy typically occurs when a nonunitary operator $\mathcal{V}$
is available that provides a high value of fidelity.
An example of such an operator can be a solution to the optimization problem (\ref{FUMax})
subject to partial constraints (Eq. (\ref{KrausPartialConstraint}) with $N_s=1$),
which does not satisfy the full set of constraints (\ref{FUMaxConstraints}).
The expansion (\ref{wsExpr}) allows constructing a mixed unitary channel that provides
a similar value of fidelity. The problem of constructing a quantum channel is reduced to two steps:
    1. Find a nonunitary operator to estimate the maximum possible fidelity, and then
    2. Find a mixed unitary channel that gives approximately the same fidelity.
This approach allows us to consider a sequence of $N_s = 1$ problems instead of a single problem with a high Kraus rank $N_s$.
The only factor that limits this program is the formulation of the actual objective function as a quadratic form on quantum channel operators.
In most cases, this is possible only through some kind of approximation, which significantly limits its applicability.

\subsection{\label{DemoHierarchy}A Demonstration of Constructed Hierarchy of Operators}
Let us  demonstrate the construction of operators hierarchy.
First,
we would like to mention that this problem is more
difficult for numerical solution than the original (\ref{FUMax}).
With the ``external'' constraints
(\ref{constraintsUPreviousOrthogonalChannel})
added, we were able to solve the problem with $N_s$ not greater than $5$
to $10$, depending on the values of $D$ and $n$.
The first difficulty arises from numerical instability
and
the need for a new 
\hyperref[Cexplicit]{adjustment algorithm}
to satisfy both the (\ref{FUMaxConstraints}) and (\ref{constraintsUPreviousOrthogonalChannel}) constraints, however
this problem is technical and can be resolved with a little effort.
The second difficulty is a significant one.
The optimization problem we are able to solve is a quadratic functional 
(\ref{FasOperatorsProduct}) optimization. However,
mixed states fidelity
(\ref{fidelityStdDefinition})
is not of this form, thus it cannot be directly applied to a general quantum channel.
In the unitary case where $N_s=1$, there are several good  options,
the most noticeable being $F^{\sqrt{\varrho}\sqrt{\rho}}$ (\ref{FSQRT}).
In cases where $N_s>1$, there are no good options, as
depicted in Fig.\ref{SkSQRTFig}b.
For this reason we try a number of ``proxies'' for total fidelity (\ref{FQuantumChannelMapping}).
Specifically
$F^{\varrho\sigma}$ (\ref{QpureStateDensityMatrix}),
$F^{v}$  (\ref{fidelityScalProduct}),
$F^{N \varrho ^2}$ (\ref{QpureStateDensityMatrixNormalized})
and $F^{\sqrt{\varrho}\sqrt{\rho}}$ (\ref{FSQRT}).
These four expressions are quadratic in $\mathcal{U}$ what allows
us to build a $S_{jk;j^{\prime}k^{\prime}}$ to which the optimization technique
can be applied.

Consider a demonstration. We generate a quantum channel (\ref{KrausOperator})
of dimension $D=n=10$
with Kraus rank $N_s=3$. Then generate $M=1000$ random states
$\rho^{(l)}_{kk^{\prime}}$ (\ref{randomRho}) of rank $N_r=1$ (pure states).
Passing them through the quantum channel obtain
the $\varrho^{(l)}_{jj^{\prime}}$ of rank $3$.
If we calculate the total fidelity (\ref{FQuantumChannelMapping})
with the state closeness $F^{prop}$ (\ref{fidelityStdDefinition})
of this $\rho^{(l)}_{kk^{\prime}}\to\varrho^{(l)}_{jj^{\prime}}$
mapping on $B_s$ of the quantum channel, then the result will be
$\mathcal{F}^{prop}(B^{exact}_s)=M=1000$ since these $B_s$ make an exact mapping
of the density matrix for all $l$.
However, $F^{prop}$ is not a quadratic function on $B_s$
because of the square root taken from the sum (\ref{KrausOperator})
--- our optimization algorithm cannot be applied.
Instead, for the four aforementioned fidelity proxies
that produce a quadratic target functional (\ref{FasOperatorsProduct}),
we calculate a superoperator $S_{jk;j^{\prime}k^{\prime}}$
from density matrix mapping,
perform optimization,
and construct $N_s=3$
unitary hierarchy. The result is presented
in Table \ref{tabHierarachy}, see
\texttt{\seqsplit{com/polytechnik/algorithms/DemoDMGeneralMappingTest.java}}
for an implementation and run it as
\texttt{\seqsplit{java\ com/polytechnik/algorithms/DemoDMGeneralMappingTest\ 2>\&1\ |\ grep\ Proxy=}}.

\begin{table*}
\caption{\label{tabHierarachy}
A demonstration of a unitary hierarchy for a quantum channel of dimension $D=n=10$
with Kraus rank $N_s=3$.
A total of $M=1000$ random density matrices
$\rho^{(l)}_{kk^{\prime}}$ (\ref{randomRho}),
each of rank $N_r=1$ (pure states),
are generated and mapped to $\varrho^{(l)}_{jj^{\prime}}$
using the quantum channel with Kraus rank $N_s=3$.
The mapping 
gives $\mathcal{F}^{prop}(B^{exact}_s)=M=1000$
for proper closeness (\ref{fidelityStdDefinition}),
but we need a quadratic expression (\ref{FasOperatorsProduct})
for total fidelity. Four fidelity proxies are considered:
$F^{\varrho\sigma}$ (\ref{QpureStateDensityMatrix}),
$F^{v}$  (\ref{fidelityScalProduct}),
$F^{N \varrho ^2}$ (\ref{QpureStateDensityMatrixNormalized})
and $F^{\sqrt{\varrho}\sqrt{\rho}}$ (\ref{FSQRT})
to construct four different hierarchies.
For $N_s>1$,
there always exists a single unitary mapping that produces a proxy-fidelity greater than that of the entire quantum channel,
e.g. for $F^{\sqrt{\varrho}\sqrt{\rho}}$
we have $\mathcal{F}(B^{exact}_s)=617.13<\mathcal{F}^{[0]}=627.04$,
whereas $\mathcal{F}^{prop}(B^{exact}_s)=1000>\mathcal{F}^{prop}(\mathcal{U}^{[0]})=627.04$.
Also, note that among all four proxies,
$\mathcal{F}^{prop}(\mathcal{U}^{[i]})$
is exactly equal to $\mathcal{F}^{[i]}$ only for the proxy-fidelity $F^{\sqrt{\varrho}\sqrt{\rho}}$,
this follows from matching (\ref{FSQRT}) to (\ref{fidelityStdDefinition}) in the case $N_s = 1$.
}
\begin{ruledtabular}
\npdecimalsign{.}
\nprounddigits{2}
\npfourdigitnosep
\begin{tabular}{l|n{5}{2}|n{5}{2}n{5}{2}n{5}{2}n{5}{2}n{5}{2}n{5}{2}}
\multicolumn{1}{l|}{$F$ constr.} &
\multicolumn{1}{c|}{$\mathcal{F}(B^{exact}_s)$} &
\multicolumn{1}{c}{$\mathcal{F}^{[0]}$} &
\multicolumn{1}{c}{$\mathcal{F}^{prop}(\mathcal{U}^{[0]})$} &
\multicolumn{1}{c}{$\mathcal{F}^{[1]}$} &
\multicolumn{1}{c}{$\mathcal{F}^{prop}(\mathcal{U}^{[1]})$} &
\multicolumn{1}{c}{$\mathcal{F}^{[2]}$} &
\multicolumn{1}{c}{$\mathcal{F}^{prop}(\mathcal{U}^{[2]})$}
\\
\hline
$F^{\varrho\sigma}$ &
393.83174056666746 &
410.59157706919996 & 625.0040306604695 &
344.93589966221094 & 558.09695974508 &
303.1652344494753 & 507.43668942724173 \\
$F^{v}$ &
626.7984497731115 &
652.4722048564872 & 625.1764439455511 &
552.3639263428842 & 559.8023064311709 &
487.66717379992394 & 510.978866576248 \\
$F^{N \varrho ^2}$ &
999.9999999999999 &
1039.4442877993176 & 625.3343362584491 &
886.3757723357511 & 561.4511329086889 &
786.06798160551 & 514.4057751611593 \\
$F^{\sqrt{\varrho}\sqrt{\rho}}$ &
617.1272194016092 &
627.044255728697 & 627.04425541654 &
571.8589158842116 & 571.8589162721518 &
545.9414912987214 & 545.9414914815213
\end{tabular}
\end{ruledtabular}
\end{table*}

First, we would note that all proxies (except $F^{N \varrho ^2}$)
do not give $M$ on exact quantum channel mapping $\mathcal{F}(B^{exact}_s)$;
but this may be fixed by normalizing.
The major difficulty is the fact that for all four proxies
the maximal total proxy-fidelity is reached not on
the exact mapping $B_s$ of quantum channel.
See the column $\mathcal{F}^{[0]}$ ---
the total proxy-fidelity calculated on a single unitary operator
$\mathcal{U}^{[0]}$ exceeds the proxy-fidelity $\mathcal{F}(B^{exact}_s)$
of the entire quantum channel $B_s$.
This is a general property of all quadratic on $B_s$
fidelity forms (\ref{FasOperatorsProduct}) we have tested,
the four listed in Table \ref{tabHierarachy}
as well as a few others: There exists a single unitary operator,
the $\mathcal{U}^{[0]}$ in (\ref{FexpansionOrt}) hierarchy,
that yields proxy-fidelity exceeding that of the entire quantum channel.
This is the primary issue encountered when applying our
QCQP algorithm to reconstruct a quantum channel with a Kraus rank $N_s>1$.
Whereas for $N_s=1$ density matrix unitary mapping we have a number of
good options for quadratic on $\mathcal{U}$ fidelity,
in the case of $N_s>1$,
for all proxy-fidelities we have tested,
there always exists a single unitary mapping that yields a proxy-fidelity greater than that of the entire quantum channel.

A quantum channel is always constructed with a preset mapping.
The constraints (\ref{constraintKrauss})
mean that
a unit matrix from the Hilbert space 
\emph{IN}
should be mapped
to a unit matrix
in the Hilbert space \emph{OUT}.
In \cite{malyshkin2022machine,belov2024partiallyPRE} a quantum channel
converting the Gram matrix in \emph{IN} into the Gram matrix
in \emph{OUT} was considered, with a regularization
it is equivalent to the same (\ref{constraintKrauss}).
In Section \ref{UnitaryMappingOfMixedStates}
a quantum channel mapping 
$\sqrt{\rho}\to\sqrt{\varrho}$ was considered,
in the $N_s=1$ unitary case the same quantum channel maps
both $\rho$ and $\sqrt{\rho}$ what allowed us to apply unitary learning
to density matrix mappings. The problem of constructing
a good quadratic proxy-fidelity
in the case of $N_s>1$ is a subject for future research.

\subsection{\label{TracePreservingMaps}Trace Preserving Maps}
A general quantum channel (\ref{KrausOperator}) is a mapping between
two Hilbert spaces. In this paper, we primarily use it to convert
the density matrix.
Such a conversion may not preserve the matrix trace.
For example, a partially unitary mapping with $D<n$ and $N_s=1$ was considered in \cite{belov2024partiallyPRE}.
Mappings that preserve the matrix trace have
\href{https://learning.quantum.ibm.com/course/general-formulation-of-quantum-information/quantum-channels#channels-transform-density-matrices-into-density-matrices}{special importance}
in quantum channel studies.
A question arises: when does a general quantum channel preserve the matrix trace?
We are interested in formulating the optimization
problem (\ref{KrausMax}) with quadratic constraints, e.g. (\ref{KrausOrt1}).
Consider a density matrix $\rho_{kk^{\prime}}$.
Applying a quantum channel with the operators $b_{s,jk}$
to it and taking the trace yields the $\mathrm{Tr} \varrho$.
\begin{align}
\mathrm{Tr}\varrho&=
\sum\limits_{s=0}^{N_s-1}
\sum\limits_{j=0}^{D-1}
\sum\limits_{k,k^{\prime}=0}^{n-1}
b_{s,jk}\rho_{kk^{\prime}}b^*_{s,jk^{\prime}}
\label{varrhoTransfTrace}
\end{align}
For the expression to ensure $\mathrm{Tr}\varrho=\mathrm{Tr}\rho$
for an arbitrary $\rho$, these conditions should be satisfied:
\begin{align}
\delta_{kk^{\prime}}&=
\sum\limits_{s=0}^{N_s-1}
\sum\limits_{j=0}^{D-1}
b_{s,jk}b^*_{s,jk^{\prime}}
\label{TracePreservingConstraint}
\end{align}
Equation (\ref{TracePreservingConstraint}) is a familiar trace preservation condition
\begin{align}
  \sum\limits_{s=0}^{N_s-1} B_s^{\dagger}B_s&=\mathds{1} \label{constraintKraussSpur}
\end{align}
There are $n(n+1)/2$
independent constraints on the quantum channel (\ref{KrausOperator}) to
\hyperref[condTraceFootnote]{preserve} the trace. 
For $N_s=1$ it immediately gives $n=D$.
For $n=D$  and an arbitrary $N_s$, 
(\ref{constraintKrauss}) and (\ref{constraintKraussSpur})
are identical.
In the convex combination of unitary channels gauge (\ref{UnitaryHierarchyQC}) with $D=n$
it gives (\ref{wNorm1}).
For trace calculating  map (where $D=1$) it gives $N_s=n$.
For these trace-preserving constraints
their
linear forms (\ref{TracePreserveCKK}),
totaling $(n-1)(n+2)/2$,
can be
\hyperref[optionalConstraintsTracePreserve]{added}
to the constraints (\ref{KrauslinearConstraintsHomog})
of the numerical method from Appendix \ref{numericalSolutionSect}.

For $n=D$, the constraints (\ref{TracePreservingConstraint})
are equivalent to (\ref{KrausOrt1}).
For $D<n$ the latter has fewer constraints than the former.
When $D<n$, the value of $N_s$
should be chosen to be large enough
so that the Gram matrix (\ref{GramKrausKK}) is not degenerate,
see (\ref{traceCalculatingB}) as an example of
a trace-calculating quantum channel having $D=1$ and $N_s=n$.
For $D<n$ the matrix $B_s$ has dimensions $D\times n$,
which gives the matrix $B_s^{\dagger}B_s$ a rank of $D$.
Thus, for $D<n$, in the case of trace-preserving maps,
the Kraus rank should be at least
\begin{align}
N_s&\ge n-D+1
\label{minKrausRank}
\end{align}
which is the minimum Kraus rank.
Otherwise, the Gram matrix (\ref{GramKrausKK}) becomes degenerate,
and the trace preservation condition (\ref{constraintKraussSpur}) cannot be satisfied.

\section{\label{conclusion}Conclusion}

We construct ``quantum mechanics over quantum mechanics''
by generalizing eigenstates to quantum channel mapping operators.
While in traditional quantum mechanics
the stationary Schr\"{o}dinger equation
determines system eigenstates,
in our approach the algebraic problem (\ref{eigenvaluesLikeProblem})
determines quantum channel mapping operators.
The total
fidelity must be a quadratic function on operators (\ref{FasOperatorsProduct})
for the problem to be represented in the form (\ref{eigenvaluesLikeProblem}).

The technique was applied to the unitary mapping
of density matrices.
Since, for unitary mapping, the same quantum channel
converts both the density matrix and its square root,
the most promising approach is to convert the density matrix mapping
(\ref{mlproblemVectorDensityMatrix})
to the density matrix square root mapping
(\ref{mlproblemVectorDensityMatrixSQRT}),
to which we can apply our QCQP optimization algorithm.
This allows us to employ unitary learning in the application to density matrix mapping, representing an important advancement from the commonly studied unitary mapping of pure states
$\phi_l=\mathcal{U} \psi_l$;
it allows us to distinguish between a probabilistic mixture of states and their superposition.
The technique was tested on a number of randomly generated
density matrices $\rho^{(l)}$ of different ranks
unitary mapped to $\varrho^{(l)}$; in all cases
the unitary quantum channel (\ref{operatorTransform})
was perfectly recovered.

The problem was then generalized to quantum channels (\ref{KrausOperator}) with Kraus rank $N_s>1$.
Reconstructing a general quantum channel mapping appears to be significantly more challenging.
First, we were unable to represent the proper fidelity (\ref{fidelityStdDefinition})
as a quadratic form of the mapping operators,
necessitating the use of approximations discussed in Section \ref{formulationOfTheProblem}.
Developing a better quadratic fidelity representation (for mapping operators) is a subject for future research and
is closely tied to the physical interpretation of quantum channels.
Second, even when an approximate quadratic form for fidelity is obtained, the algorithm described
in Appendix \ref{numericalSolutionSect} fails to converge for $N_s>1$.
We expect, however, that implementing advanced constraint methods could lead to improvements.
If, instead of reconstructing an arbitrary quantum channel,
we restrict the problem to constructing a mixed unitary channel (\ref{UnitaryHierarchyQC})
using a hierarchy of unitary operators, then the problem becomes solvable, as detailed in Section \ref{Kraushierarchy}.
The results, however, are less satisfactory than desired due to a ``double approximation'': first,
approximating fidelity with a quadratic form,
and second, relying on a mixed unitary quantum channel instead of a general one.

In this work, we have studied quantum channels that convert:
a unit matrix from \emph{IN} to \emph{OUT} (standard definition),
a Gram matrix from \emph{IN} to \emph{OUT},
and a $\sqrt{\rho}$ to $\sqrt{\varrho}$.
We developed a method for finding the global maximum by solving
a novel algebraic problem (\ref{eigenvaluesLikeProblem}),
see Appendix \ref{SchodingerNonStationary},
which presents a generalization of this algebraic problem
to the nonstationary case and introduces a time-dependent Schr\"{o}dinger-like equation
for operator $\mathcal{U}(t)$.
We anticipate that applying memoryless quantum channel mappings
to other problems could provide a solid foundation for a new form of machine learning knowledge representation.
For initial insights into quantum channels with memory, see Appendix \ref{StatesWithMemory} below.

\begin{acknowledgments}
This research was supported by Autretech Group,
\href{https://xn--80akau1alc.xn--p1ai/}{www.{\fontencoding{T2A}\fontfamily{cmr}\selectfont атретек.рф}},
a resident company of the Skolkovo Technopark.
  We thank our colleagues from the Autretech R\&D department
  who provided insight and expertise that greatly assisted the research.
  Our grateful thanks are also extended
  to Mr. Gennady Belov for his methodological support in doing the data analysis.

In this work, we study quantum channels as a positive map between two Hilbert spaces. In the late 80s to early 90s, Ivan Anatol'evich Komarchev brought to V.M.'s attention the problem of positive maps between Hilbert spaces and their relation to the Radon-Nikodym derivative. This became the origin of this entire theme. His unexpected death in 2022 was a significant loss to the Department of Mathematics at
\href{https://english.spbstu.ru/}{St.Petersburg Polytechnic University},
his colleagues who collaborated with him, and everyone who knew Ivan Anatol'evich.
This work is dedicated to his memory.
\end{acknowledgments}

\appendix
\section{\label{numericalSolutionSect}Numerical Solution}

The described numerical algorithm is a further development
of the one presented in \cite{belov2024partiallyPRE}.
The advancements over the previous version include:
\begin{itemize}
\item The ability to simultaneously include quadratic constraints of different types,
such as (\ref{KrausOrt1}) and (\ref{KrausCanonic}),
or alternatively, (\ref{TracePreservingConstraint}) and (\ref{KrausCanonic}),
requires the introduction of two kinds of Lagrange multipliers: $\lambda_{ij}$ and $\nu_{ss^{\prime}}$.
These multipliers are obtained from the solution of the linear system (\ref{sumSQL}).

\item
Different groups of quadratic constraints create different convergence-helper linear constraints.
A general method, which does not involve the use of a special basis as in \cite{belov2024partiallyPRE},
is formulated in Appendix \ref{LinearConstraints}. This method involves varying the total
$N_c$ quadratic constraints to obtain $N_c-1$ homogeneous linear constraints
and a single ``simplified'' (partial) quadratic constraint (\ref{KrausPartialConstraint}),
which facilitates the use of a standard eigenvalue problem solver as an iterative algorithm building block.

\item
In addition to the convergence-helper constraints, we can now include ``external'' linear constraints.
These external constraints can be used to construct a hierarchy of solutions for the novel algebraic problem (\ref{eigenvaluesLikeProblem}).

\item
Now that we have several types of constraints, especially ``external'' and convergence-helper constraints,
the procedure for adjusting the solution to satisfy all constraints becomes more complicated.
This is because adjustments for different groups of constraints may conflict with each other.
Therefore, a conflict resolution strategy is required, such as the one discussed in Appendix \ref{KrausAdjustOrthogonal}.

\item
The algorithm described below converges only when $N_s=1$: maximize (\ref{FUMax})
subject to the constraints (\ref{FUMaxConstraints}).
Currently, it does not work for a general quantum channel where $N_s>1$,
as the problem becomes degenerate and the algorithm fails to converge.
However, the calculations below are presented for an arbitrary $N_s$
because the technique provides a clear approach for adding constraints of various types.
The current version of the algorithm converges only for unitary (and partially unitary) learning when $N_s=1$,
with the ``external'' constraints possibly used to build a \hyperref[Kraushierarchy]{hierarchy of unitary operators}.

\end{itemize}

Consider the optimization problem: optimize (\ref{KrausMax})
subject to (\ref{KrausOrt1})
and the canonical gauge (\ref{KrausCanonic})
constraints.
For a general $N_s$ the solution
to the optimization problem 
is a matrix
$b_{s,jk}$ of dimensions $N_s\times D\times n$,
corresponding to $N_s$ 
Kraus operators $B_s$ (\ref{KrausOperator})
that satisfy the orthogonality constraints (\ref{constraintKrauss})
and the gauge constraints, i.e. (\ref{KrausCanonicForm}).
This is the most general form of mapping
a Hilbert space $\Ket{\psi}$ of dimension $n$
into a Hilbert space $\Ket{\phi}$ of dimension $D$.
This is a variant of the \href{https://en.wikipedia.org/wiki/Quadratically_constrained_quadratic_program}{QCQP} problem
and the technique we used in \cite{belov2024partiallyPRE} can also be applied here.
The tensor $S_{jk;j^{\prime}k^{\prime}}=S^*_{j^{\prime}k^{\prime};jk}$ is Hermitian, it does not depend on the Kraus index $s$; the calculations below
can be generalized to $s$-dependent
$S_{s,jk;s^{\prime},j^{\prime}k^{\prime}}$ except for Kraus operators transformation
to a canonical form with Appendix \ref{KrausCanonicMethod}.
For simplicity we consider $S_{jk;j^{\prime}k^{\prime}}$ and $b_{s,jk}$ to be real and do not write the complex conjugated $^*$ below,
a generalization to complex values is straightforward.
Consider a Lagrangian
\begin{align}
  \mathcal{L}&=
  \sum\limits_{s=0}^{N_s-1}
  \sum\limits_{j,j^{\prime}=0}^{D-1}\sum\limits_{k,k^{\prime}=0}^{n-1}
             b_{s,jk}S_{jk;j^{\prime}k^{\prime}}b_{s,j^{\prime}k^{\prime}} \nonumber \\
             &+
   \sum\limits_{s=0}^{N_s-1}\sum\limits_{j,j^{\prime}=0}^{D-1}        
   \lambda_{jj^{\prime}}\left[\delta_{jj^{\prime}}-\sum\limits_{k^{\prime}=0}^{n-1}b_{s,jk^{\prime}} b_{s,j^{\prime}k^{\prime}} \right] \nonumber \\
   &-
   \sum\limits_{s\ne s^{\prime}=0}^{N_s-1}
   \sum\limits_{j=0}^{D-1}\sum\limits_{k=0}^{n-1}
   \nu_{ss^{\prime}} b_{s,jk} b_{s^{\prime},jk}
   \xrightarrow[b]{\quad }\max
   \label{KrausLagrangian}
\end{align}
There are $D(D+1)/2$ constraints (\ref{KrausOrt1})
and $N_s(N_s-1)$ constraints (\ref{KrausCanonic}).
Lagrange multipliers $\lambda_{jj^{\prime}}$ and $\nu_{ss^{\prime}}$
are Hermitian matrices with corresponding number of independent elements;
the $\nu_{ss^{\prime}}$ matrix has all diagonal elements equal to zero.
Introduce a matrix $\mathcal{S}_{jk;j^{\prime}k^{\prime}}$
to consider
the quadratic form maximization problem
obtained with the partial
constraint (\ref{KrausPartialConstraint})
\begin{align}
  &\mathcal{S}_{s,jk;s^{\prime},j^{\prime}k^{\prime}}=
             \delta_{ss^{\prime}}S_{jk;j^{\prime}k^{\prime}}
             -\lambda_{jj^{\prime}}\delta_{ss^{\prime}}\delta_{kk^{\prime}}
             -\nu_{ss^{\prime}} \delta_{jj^{\prime}}\delta_{kk^{\prime}}
   \label{KrausSwithLagrange}\\
&
\frac{
\sum\limits_{s,s^{\prime}=0}^{N_s-1}
\sum\limits_{j,j^{\prime}=0}^{D-1}\sum\limits_{k,k^{\prime}=0}^{n-1}
             b_{s,jk}\mathcal{S}_{s,jk;s^{\prime},j^{\prime}k^{\prime}}b_{s^{\prime},j^{\prime}k^{\prime}}
             }
             {\frac{1}{D}
             \sum\limits_{s=0}^{N_s-1}
\sum\limits_{j=0}^{D-1}
\sum\limits_{k=0}^{n-1}b^2_{s,jk}
             }
   \xrightarrow[b]{\quad }\max
   \label{KrausOptimizationVarS}
\end{align}
Following \cite{belov2024partiallyPRE}, we calculate Lagrange
multipliers $\lambda_{jj^{\prime}}$ and $\nu_{ss^{\prime}}$
from the current iteration $b_{s,jk}$,
see Appendix \ref{lagrangeMultipliersCalculation} below ---
then $\mathcal{S}_{s,jk;s^{\prime},j^{\prime}k^{\prime}}$ is fixed
and the problem (\ref{KrausOptimizationVarS})
can be considered as an eigenvalue problem.
Additional $N_d$ linear constraints on $b_{s,jk}$ can be incorporated
to improve convergence
or obtain a sequence of solutions,
see Appendix \ref{LinearConstraints} for convergence improving
constraints and (\ref{constraintsUPreviousOrthogonalChannel}) for obtaining a solution sequence.
\begin{align}
0&=\sum\limits_{s=0}^{N_s-1}\sum\limits_{j=0}^{D-1}\sum\limits_{k=0}^{n-1} 
C_{d;s,jk}b_{s,jk}
 & d=0\dots N_d-1
\label{KrauslinearConstraintsHomog}
\end{align}
A common method of solving the eigenproblem (\ref{KrausOptimizationVarS})
with
\href{https://en.wikipedia.org/wiki/System_of_linear_equations#Homogeneous_systems}{homogeneous} linear constraints (\ref{KrauslinearConstraintsHomog})
is Lagrange multipliers method\cite{golub1973some}.
Since we have a large number of constraints
it is better to use direct elimination instead.
From $N_sDn$ independent components of $b_{s,jk}$ select
some general variables $V_p$, $p=0\dots N_V-1$
\begin{align}
N_V&=N_sDn-\mathrm{rank}(C_{d;s,jk}) \label{KrausNvvariables} \\
b_{s,jk}&=\sum\limits_{p=0}^{N_V-1}M_{s,jk;p} V_p
\label{KrausuExprVp}
\end{align}
where the constraints in (\ref{KrauslinearConstraintsHomog}) are eliminated.
The method of selection could be
\href{https://en.wikipedia.org/wiki/Gaussian_elimination}{Gaussian elimination},
\href{https://en.wikipedia.org/wiki/QR_decomposition}{QR decomposition},
or a similar technique.
A simple implementation with row and column pivoting
is used in
\texttt{\seqsplit{com/polytechnik/utils/EliminateLinearConstraints\_HomegrownLUFactorization.java}}.
The result is a matrix $M_{s,jk;p}$ that converts
$N_V$ independent variables $V_p$
to $b_{s,jk}$ of $N_sDn$ components satisfying
all the constraints (\ref{KrauslinearConstraintsHomog}).
A new eigenproblem has the matrices $\mathcal{S}_{p;p^{\prime}}$
and $Q_{p;p^{\prime}}$  in the numerator and denominator respectively.
\begin{subequations}
\label{KrausmatrTransf}
\begin{align}
\mathcal{S}_{p;p^{\prime}}&=
\sum\limits_{s,s^{\prime}=0}^{N_s-1}
\sum\limits_{j,j^{\prime}=0}^{D-1}\sum\limits_{k,k^{\prime}=0}^{n-1}
M_{s,jk;p}
\mathcal{S}_{s,jk;s^{\prime},j^{\prime}k^{\prime}}
M_{s^{\prime},j^{\prime}k^{\prime};p^{\prime}} \label{KraustransfS} \\
Q_{p;p^{\prime}}&=
\sum\limits_{s=0}^{N_s-1}
\sum\limits_{j=0}^{D-1}\sum\limits_{k=0}^{n-1}
M_{s,jk;p}
M_{s,jk;p^{\prime}} \label{KraustransfQ}
\end{align}
\end{subequations}
Then we can write the eigenproblem (\ref{KrausOptimizationVarS}) in the form,
\begin{align}
&\frac{
\sum\limits_{p,p^{\prime}=0}^{N_V-1}
V_p
\mathcal{S}_{p;p^{\prime}}
V_{p^{\prime}}
}{
\sum\limits_{p,p^{\prime}=0}^{N_V-1}
V_pQ_{p;p^{\prime}}V_{p^{\prime}}
} 
 \xrightarrow[V]{\quad }\max
 \label{KrausEPLV} \\
  &   \sum\limits_{p^{\prime}=0}^{N_V-1}
\mathcal{S}_{p;p^{\prime}}V^{[i]}_{p^{\prime}}
=\mu^{[i]}\sum\limits_{p^{\prime}=0}^{N_V-1} Q_{p;p^{\prime}} V^{[i]}_{p^{\prime}}
\label{KrausEPLeVV}
\end{align}
which has $\mathrm{rank}(C_{d;s,jk})$ less dimension and no linear constraints.
The matrix $Q_{p;p^{\prime}}$ in the denominator is no longer
a unit matrix.
This is not an issue since
any modern linear algebra package internally converts
a generalized eigenproblem to a regular one,
see e.g.
\href{https://www.netlib.org/lapack/lug/node54.html}{DSYGST, DSPGST, DPBSTF}
and similar subroutines.
Combining all together the algorithm becomes:
\begin{enumerate}
\item
\label{firstStepLambda}
  Take initial $\lambda_{ij}$, $\nu_{ss^{\prime}}$
  and linear constraints $C_{d;s,jk}$
to solve the optimization problem (\ref{KrausEPLV})
  with respect to $V_p$.
 The solution method involves solving an eigenvalue problem of dimension
$N_V$,
the number of columns in $M_{s,jk;p}$ matrix.
A new $b_{s,jk}$ is obtained from $V_p$ using (\ref{KrausuExprVp}).
The result: $i=0\dots N_V-1$ eigenvalues $\mu^{[i]}$ and corresponding
matrices $b_{s,jk}^{[i]}$ reconstructed from $V_p^{[i]}$.
The value of $N_V$ is typically $N_sDn-(D-1)(D+2)/2-N_s(N_s-1)/2$.
Additional constraints (further reducing $N_V$) are added when
constructing the operators hierarchy.
\item
\label{EVSelectionFromAll}
A heuristic is required
to select the $b_{s,jk}$  among all $N_V$ eigenstates.
There is a discussion  about this in \cite{belov2024partiallyPRE}.
Our numerical experiments show that in most cases it is sufficient
to always take the state of the maximal $\mu^{[i]}$
to reach the global maximum.
In the current implementation we always 
select this state.
\item \label{lagrangeMultipliersStep}
Obtained $b_{s,jk}$
satisfies only partial constraint (\ref{KrausPartialConstraint}).
We need to adjust it to satisfy all the required constraints
of orthogonality and canonical form.
Adjust $b_{s,jk}$
with Appendix \ref{KrausAdjustOrthogonal} to satisfy
orthogonality (\ref{KrausOrt1}),
then apply the result to Appendix \ref{KrausCanonicMethod}
to convert the Kraus operators to canonical form (\ref{KrausCanonic}).
For $N_s=1$, transformation to the canonical form is not required.
With the resulting $b_{s,jk}$ 
calculate Lagrange multipliers
$\lambda_{ij}$ and $\nu_{ss^{\prime}}$
for the next iteration
as described in Appendix \ref{lagrangeMultipliersCalculation}.

\item
For good convergence, in addition to Lagrange multipliers,
we need to select a subspace for the next iteration variation.
Using full size basis leads to poor convergence \cite{malyshkin2022machine}.
Use Appendix \ref{LinearConstraints}
to obtain
(from $b_{s,jk}$)
the linear constraints coefficients 
$C_{d;s,jk}$ used in (\ref{KrauslinearConstraintsHomog}).
There is an important feature of this additional linear constraints
approach. In addition to the constraints
from Appendix \ref{LinearConstraints},
we can manually add some external constraints,
such as requiring zero projection
(\ref{constraintsUPreviousOrthogonalChannel})
onto already found solutions.
This way
a hierarchy of high $\mathcal{F}$ solutions can be obtained.
\item             
    Put these new
$\lambda_{ij}$ and $\nu_{ss^{\prime}}$
into (\ref{KrausSwithLagrange})
    and, using the basis $V_p$ obtained (\ref{KrausuExprVp}) from $C_{d;s,jk}$,
    calculate the generalized eigenproblem numerator and denominator matrices (\ref{KrausmatrTransf})
    to be used in the next iteration.     
    Repeat the iteration process until convergence
    to a maximum (presumably global) of $\mathcal{F}$ with $b_{s,jk}$ satisfying
    the constraints (\ref{KrausOrt1}) and (\ref{KrausCanonic}).
    If convergence is achieved, the Lagrange multipliers stop changing
    from iteration to iteration, and the $\mu^{[i]}$ of the selected state
    in step \ref{EVSelectionFromAll}
    is close to zero.
    On the first iteration take zero initial values for the Lagrange multipliers and have no linear constraints.
\end{enumerate}
For a general $N_s$ implementation of this algorithm
see
\texttt{\seqsplit{com/polytechnik/kgo/IterationalSubspaceLinearConstraintsNaiveKraus.java}}.
There is a special $N_s=1$ implementation of it
\texttt{\seqsplit{com/polytechnik/kgo/KGOIterationalSubspaceLinearConstraintsB.java}}; they share a codebase and unit tests.
Currently the algorithm converges only if $N_s=1$;
in this case,
except for implementation improvements and optimizations,
it has properties very similar to the
\texttt{\seqsplit{com/polytechnik/kgo/KGOIterationalSubspaceLinearConstraints.java}}
implementation in \cite{belov2024partiallyPRE}.

\subsection{\label{KrausCanonicMethod}Transforming Kraus Operators to Canonical Form}
The selection of Kraus operators $B_s$ in which to evaluate (\ref{KrausMax}) is nonunique,
there should be a gauge
that regulates redundant degrees of freedom.
For example consider data of exact unitary mapping
$\Ket{\psi_l}\Bra{\psi_l} \to \Ket{\phi_l}\Bra{\phi_l}$
with an operator $\mathcal{U}$
providing perfect coverage in (\ref{allProjUKxfAppendix}).
Let we want to describe this data with (\ref{KrausOperator}) mapping
of Kraus rank $N_s=2$. Then any combination
$B_s=w_s \mathcal{U}$ with $1=\sum_s|w_s|^2$ is a solution.
The problem becomes degenerate. This creates difficulty in both analyzing the results and achieving convergence in the iterative algorithm -- any degeneracy greatly degrades it.
It is common to consider Kraus operators $B_s$
in the canonical form (\ref{KrausCanonicForm}).
The problem becomes:
converting a given $b_{s,jk}$ matrix to the canonical form
of Eq. (\ref{KrausCanonic}).

Kraus operators can be considered as some superoperator states,
similar 
to density matrix states,
but ket $\Ket{\cdot}$ and bra $\Bra{\cdot}$ are now operators:
$\sum_s\Ket{B_s}\Bra{B_s}$,
not vectors.
This algebra
can be uses
in the expansion of tensors,
such as $S_{jk;j^{\prime}k^{\prime}}$ (\ref{expansionSinUs}) ---
then evaluating $\mathcal{F}$ is an inner product
of superoperators.
In this work we are not going to discuss the construction
of superoperators algebra and its applications.
The most important application is the construction
of operators hierarchy in Section \ref{Kraushierarchy},
what allows us
to reconstruct a quantum channel from observable data.
There are a number of other
interesting features that arise on this path and
we hope to present a detailed discussion elsewhere. 
In this appendix we limit ourselves
to one small but important problem.

Assume we obtained a partially Kraus operator
in the form of a matrix
$b_{s,jk}$ that evaluates the fidelity to some $\mathcal{F}$ (\ref{KrausMax}).
This $b_{s,jk}$ does not satisfy the (\ref{KrausCanonic}) constraints.
The problem is to convert $b_{s,jk}$ to a new $\widetilde{b}_{s,jk}$
satisfying the (\ref{KrausCanonic})
while maintaining the same value of the functional
 $\mathcal{F}$. The solution
is not unique and requires a gauge, akin to basis orthogonalization.
Since the tensor $S_{jk;j^{\prime}k^{\prime}}$ does not depend on the Kraus index $s$,
a solution can be readily found.
We use the same technique previously employed
in adjusting for partial unitarity\cite{belov2024partiallyPRE},
as outlined in Appendix \ref{KrausAdjustOrthogonal}.
It's even simpler to apply it here since diagonal elements do not enter into the constraints defined in (\ref{KrausCanonic}).
Consider the Gram matrix in the Kraus space
\begin{align}
G_{ss^{\prime}}&=\sum\limits_{j=0}^{D-1}\sum\limits_{k=0}^{n-1}b_{s,jk}b_{s^{\prime},jk}
\label{GramUpartialKraus}
\end{align}
and solve an eigenproblem with it
\begin{align}
\Ket{G\middle|g^{[i]}}&=\lambda_G^{[i]}\Ket{g^{[i]}} &i=0\dots N_s-1
\label{GramMatrixEVKraus}
\end{align}
Since the  $S_{jk;j^{\prime}k^{\prime}}$ does not depend on $s$
one can verify that
\begin{align}
\widetilde{b}_{s,jk}&=
\sum\limits_{s^{\prime}=0}^{N_s-1} g^{[s]}_{s^{\prime}} b_{s^{\prime},jk}
\label{bKrausCanonic}
\end{align}
satisfies the (\ref{KrausCanonic}) constraints, whereas the
value of  $\mathcal{F}$ (\ref{KrausMax}) stays the same.
See \texttt{\seqsplit{com/polytechnik/kgo/TransformToCanonicalFormKraus.java}}
for an implementation. The implementation is straightforward:
calculate the Gram matrix (\ref{GramUpartialKraus}), solve
the eigenproblem (\ref{GramMatrixEVKraus}), then convert 
the original $b_{s,jk}$ to the basis of the found eigenvectors (\ref{bKrausCanonic});
$\mathcal{F}$ does not change with this transformation.

\subsection{\label{KrausAdjustOrthogonal}An Adjustment of Operators to Orthogonal}
A $b_{s,jk}$, found at some iteration, may not satisfy
all the required constraints. In Appendix \ref{KrausCanonicMethod}
we considered canonical form constraints (\ref{KrausCanonic}),
now consider orthogonality constraints (\ref{KrausOrt1}).
It is sufficient to consider a single $u_{jk}$ since the same transform
is applied to all $b_{s,jk}$.
The Gram matrix
\begin{align}
G_{jj^{\prime}}&=
\sum\limits_{k=0}^{n-1}u_{jk}u_{j^{\prime}k}
\label{KrausGramUpartialJJ}
\end{align}
(or $G_{jj^{\prime}}=\sum_{s=0}^{N_s-1}\sum_{k=0}^{n-1}b_{s,jk}b_{s,j^{\prime}k}$ for $N_s>1$)
is a unit matrix when all the constraints (\ref{KrausOrt1})
are satisfied. When this is not the  case --- we want
to change the $u_{jk}$ in a way that the change is as small
as possible. Following \cite{belov2024partiallyPRE} we apply
the same $G^{-1/2}_{jj^{\prime}}$ technique. Specifically,
solve the eigenproblem
\begin{align}
\Ket{G\middle|g^{[i]}}&=\lambda_G^{[i]}\Ket{g^{[i]}}
&i=0\dots D-1
\label{KrausGramMatrixEVJJ}
\end{align}
and calculate the inverse square root of the Gram matrix
\begin{align}
\left\|G^{-1/2}\right\|&=\sum\limits_{i=0}^{D-1}\frac{\pm 1}{\sqrt{\lambda_G^{[i]}}}
\Ket{g^{[i]}}\Bra{g^{[i]}}
\label{KrausJJmSqrtG}
\end{align}
By checking the result, one can verify that for any $u_{jk}$
producing nondegenerated Gram matrix (\ref{KrausGramUpartialJJ}) the matrix
\begin{align}
\widetilde{u}_{jk}&=
\sum\limits_{i=0}^{D-1}G^{-1/2}_{ji}u_{ik}
\label{KrausuAdjExJJ}
\end{align}
satisfies all (\ref{FUMaxConstraints}) constraints
(or $\widetilde{b}_{s,jk}=
\sum_{i=0}^{D-1}G^{-1/2}_{ji}b_{s,ik}$
and constraints (\ref{KrausOrt1})
for $N_s>1$;
in this transform we apply the same $G^{-1/2}_{ji}$
to every Kraus operator $B_s$ --- the adjustment does not depend on $s$).
Previously considered transformation  to the canonical
form
(\ref{bKrausCanonic})
has an important difference from (\ref{KrausuAdjExJJ}).
The transformation (\ref{bKrausCanonic}) does not change any observable,
it just reshuffles $B_s$ among themselves
in the (\ref{KrausOperator}) sum (gauge transform)
in a way
that
the obtained $B_s$ satisfy the canonical form constraints (\ref{KrausCanonicForm}),
specifically (\ref{KrausCanonic}).
But Eq. (\ref{KrausuAdjExJJ}) actually changes the solution,
it is not possible to satisfy (\ref{KrausOrt1}) without changing
the solution itself.
We will not repeat the lengthy discussion \cite{belov2024partiallyPRE}
about this $G^{-1/2}_{ji}$ adjustment technique,
just note that it introduces a ``minimal disturbance'' to a solution,
which is very advantageous for an iterative algorithm.
This adjustment technique
is implemented in 
\texttt{\seqsplit{com/polytechnik/kgo/AdjustToOrthogonalKraus.java}}.

There is an issue that arises when constructing operators hierarchy
in Section \ref{Kraushierarchy} --- there we do have
external linear constraints (\ref{constraintsUPreviousOrthogonalChannel})
to be added to the iterative algorithm.
Whereas the convergence-helper constraints of Appendix \ref{LinearConstraints}
are themselves calculated from the adjusted $\widetilde{u}_{jk}$ (\ref{KrausuAdjExJJ}),
external constraints
are different
 in the sense that they are preset.
The application of  $G^{-1/2}_{jj^{\prime}}$ to the current iteration
$u_{jk}$, which satisfies external constraints due to (\ref{KrausuExprVp})
incorporating all the linear constraints,
may create a $\widetilde{u}_{jk}$ that
does not satisfy external
constraints.
In most cases, this problem can be ignored since as iterations progress, the solution converges to the required subspace.
However, it would be beneficial to modify the adjustment procedure to explicitly incorporate externally defined homogeneous linear constraints.

Consider external constraints
\begin{align}
0&=
\sum\limits_{j=0}^{D-1}
\sum\limits_{k=0}^{n-1}
C^e_{d;jk}u_{jk}
\label{Cexplicit}
\end{align}
here $C^e_{d;jk}$
represents, for example, constraints 
(\ref{constraintsUPreviousOrthogonalChannel}),
where
index $d$ enumerates the constraints;
they do not include convergence-helper constraints
from Appendix \ref{LinearConstraints}, which are determined
later,
after the constraint-adjusted $u_{jk}$ is obtained.
To simplify projections bellow, consider
$C^e_{d;jk}$ and $u_{jk}$ as vectors, and convert $C^e_{d;jk}$
to an orthogonal form. One can either use Gram-Schmidt
orthogonalization or, alternatively, apply the same $G^{-1/2}$ technique
we previously
\hyperref[KrausuAdjExJJ]{used} for a different problem.
Now we convert (\ref{Cexplicit}) 
\begin{subequations}
\label{ortExplocitConstraints}
\begin{align}
G_{dd^{\prime}}&=
\sum\limits_{j=0}^{D-1}
\sum\limits_{k=0}^{n-1}C^e_{d;jk} C^e_{d^{\prime};jk}
\label{GramOrtConstr} \\
\widetilde{C}^e_{d;jk}&=
\sum_{d^{\prime}}
G^{-1/2}_{dd^{\prime}}
C^e_{d^{\prime};jk}
\label{ConstraintsOrt}
\end{align}
\end{subequations}
to an orthogonalized form
$\delta_{dd^{\prime}}= \sum_{j=0}^{D-1}
\sum_{k=0}^{n-1}\widetilde{C}^e_{d;ik} \widetilde{C}^e_{d^{\prime};ik}$;
the only purpose of this orthogonalization is to have a simple
projection formula (\ref{removeProjFromExternalConstraints}) below.
Consider the following iteration:
\begin{itemize}
\item Take the original $u_{jk}$ and convert it
to the form that satisfies orthogonality constraints
(\ref{KrausuAdjExJJ}).
\item
This $\widetilde{u}_{jk}$
satisfies  (\ref{FUMaxConstraints})
but may not satisfy external linear constraints (\ref{Cexplicit}).
Remove the projections
\begin{align}
\widetilde{\widetilde{u}}_{jk}&=
\widetilde{u}_{jk}-
\sum\limits_d
\widetilde{C}^e_{d;jk}
\sum\limits_{j^{\prime}=0}^{D-1}
\sum\limits_{k^{\prime}=0}^{n-1}\widetilde{C}^e_{d;j^{\prime}k^{\prime}}
\widetilde{u}_{j^{\prime}k^{\prime}}
\label{removeProjFromExternalConstraints}
\end{align}
Obtained $\widetilde{\widetilde{u}}_{jk}$
satisfies external constraints (\ref{Cexplicit})
but may not satisfy the orthogonality constraint (\ref{FUMaxConstraints}).
\item
Repeat the process by performing a number of iterations.
This iteration-adjustment algorithm has poor convergence per se
and requires about a hundred iterations to obtain a $u_{jk}$
that satisfies both the constraints
 (\ref{FUMaxConstraints})
and  (\ref{Cexplicit}) exactly.
However, since this iteration-adjustment
is only step \ref{lagrangeMultipliersStep}
of the main iterative algorithm, we do not necessarily need an exact solution.
Numerical experiments show that $3$ to $6$ iterations are
sufficient for the main iterative algorithm to converge
in the presence of external linear constraints.
\end{itemize}
See
\texttt{\seqsplit{com/polytechnik/kgo/AdjustToUnitaryWithEV\_SubjectToLinearConstraints.java}}
for an implementation. As emphasized above,
several iterations of these sequential adjustments are sufficient to achieve good convergence of the main algorithm. However, significant improvements are possible, and further study of this problem is necessary.

\subsection{\label{AdjustChannelToTracePreserving}An Adjustment of a Quantum Channel to Preserve the Trace}
A $b_{s,jk}$, found at some iteration, may not satisfy
the constraints (\ref{TracePreservingConstraint}) of trace preservation.
We can apply the same $G^{-1/2}$ technique from
Appendix \ref{KrausAdjustOrthogonal} to adjust the quantum channel
operators $b_{s,jk}$ to preserve the  trace.
Consider the Gram matrix
\begin{align}
G_{kk^{\prime}}&=\sum\limits_{s=0}^{N_s-1}\sum\limits_{j=0}^{D-1}b_{s,jk}b_{s,jk^{\prime}}
\label{GramKrausKK}
\end{align}
and solve an eigenproblem with it
\begin{align}
\Ket{G\middle|g^{[i]}}&=\lambda_G^{[i]}\Ket{g^{[i]}} &i=0\dots n-1
\label{GramMatrixEVKrausKK}
\end{align}
Since we typically have $D\le n$ the Gram matrix (\ref{GramKrausKK}) may
become degenerated. To avoid a degeneracy the number of terms $N_s$
in the quantum channel should be large enough,
there is a minimum Kraus rank (\ref{minKrausRank}).
For a nondegenerated $G$ calculate its inverse square root:
\begin{align}
\left\|G^{-1/2}\right\|&=\sum\limits_{i=0}^{n-1}\frac{\pm 1}{\sqrt{\lambda_G^{[i]}}}
\Ket{g^{[i]}}\Bra{g^{[i]}}
\label{KrausKKmSqrtG}
\end{align}
And apply it to the current solution $b_{s,jk}$
\begin{align}
\widetilde{b}_{s,jk}&=
\sum\limits_{q=0}^{n-1}G^{-1/2}_{kq}b_{s,jq}
\label{KrausuAdjExKK}
\end{align}
By checking the result one can verify that
$\widetilde{b}_{s,jk}$
satisfies all the trace preservation constraints (\ref{TracePreservingConstraint}),
see \texttt{\seqsplit{com/polytechnik/kgo/AdjustToTracePreservingKraus.java}}
for an implementation.
Note that the problem may become degenerate, which requires special attention. 
Also note that (\ref{KrausuAdjExKK}) is similar to (\ref{KrausuAdjExJJ})
and, contrary to (\ref{bKrausCanonic}), actually changes the $b_{s,jk}$.
A possible conflict between different adjustments
can potentially be resolved similarly to the
\hyperref[Cexplicit]{resolution}
of a conflict between linear constraints and the adjustment (\ref{KrausuAdjExJJ}).

\subsection{\label{lagrangeMultipliersCalculation}Lagrange Multipliers Calculation}
The variation of the Lagrangian $\mathcal{L}$ (\ref{KrausLagrangian})
must be zero in the iteration state $b_{s,jk}$
\begin{align}
0=\frac{1}{2}\frac{\delta\mathcal{L}}{\delta b_{s,iq}}&=
\sum\limits_{j^{\prime}=0}^{D-1}\sum\limits_{k^{\prime}=0}^{n-1}
S_{iq;j^{\prime}k^{\prime}}b_{s,j^{\prime}k^{\prime}} \nonumber \\
&- \sum\limits_{j^{\prime}=0}^{D-1} \lambda_{ij^{\prime}} b_{s,j^{\prime}q}
- \sum\limits_{s^{\prime}=0}^{N_s-1} \nu_{ss^{\prime}} b_{s^{\prime},iq}
\label{variationKrausLagrangiznZero}
\end{align}
There are a total of $N_sDn$ equations.
They are all satisfied if $b_{s,jk}$ is extremal in (\ref{KrausOptimizationVarS}).
The $b_{s,jk}$ used, however,
has the orthogonality adjustment procedure
of Appendix \ref{KrausAdjustOrthogonal} applied to it
and the Lagrangian variation is no longer zero.
Lagrange multipliers are 
Hermitian matrices $\lambda_{ij}$ and $\nu_{ss^{\prime}}$, they
have $D(D+1)/2$ and $N_s(N_s-1)/2$ independent values
thus all the $N_sDn$  equations cannot be simultaneously satisfied.
We need to select the $\lambda_{ij}$ and $\nu_{ss^{\prime}}$
to ensure they best satisfy the zero variation condition for a given $b_{s,jk}$.
Consider the $L^2$ norm of (\ref{variationKrausLagrangiznZero})
and find the $\lambda_{ij}$ and $\nu_{ss^{\prime}}$ minimizing the sum of squares.
\begin{align}
\sum\limits_{s=0}^{N_s-1}
\sum\limits_{i=0}^{D-1}
\sum\limits_{q=0}^{n-1}
\left|\frac{\delta\mathcal{L}}{\delta b_{s,iq}}\right|^2
 \xrightarrow[\lambda_{ij},\nu_{ss^{\prime}}]{\quad }\min
\label{sumSQL}
\end{align}
To simplify the minimization of (\ref{sumSQL}) it is convenient to map
$\lambda_{ij}$
into a vector of independent components $\lambda_r$.
One can note that for $j\le i$ the vector index can be taken as $r=i(i+1)/2+j$
(if $j>i$ swap them). Similarly for $\nu_r$ and $\nu_{ss^{\prime}}$
take the vector index as $r=s(s-1)/2+s^{\prime}$ for $s^{\prime}<s$ otherwise
swap the indexes. Then differentiate (\ref{sumSQL}) over $\lambda_r$ and $\nu_r$
to obtain a linear system with respect to them.
Back in \cite{belov2024partiallyPRE} a similar linear system
was analytically solved and an explicit expression for $\lambda_{ij}$
was obtained.
\begin{align}
\lambda_{ij}&=
    \mathrm{Herm}
    \sum\limits_{j^{\prime}=0}^{D-1}\sum\limits_{k,k^{\prime}=0}^{n-1}u_{ik}S_{jk;j^{\prime}k^{\prime}}u_{j^{\prime}k^{\prime}}
    \label{newLambdaSolPartial}
\end{align}
This is an analytic solution in the case $N_s=1$ with $u_{jk}=b_{0,jk}$
satisfying the (\ref{FUMaxConstraints}) constraints,
see  \texttt{\seqsplit{com/polytechnik/kgo/LagrangeMultipliersPartialSubspace.java:calculateRegularLambda}} for an implementation of this special case.
In the general case we have two types of Lagrange multipliers:
$\lambda_{ij}$ and $\nu_{ss^{\prime}}$, which prevents us from
obtaining an analytic solution. But the problem is straightforward:
take each term from (\ref{variationKrausLagrangiznZero}),
square it, and sum over all $s,i,q$. Differentiate this
sum of squares
with respect to the vectors  $\lambda_r$ and $\nu_r$ 
to obtain a linear system, then solve it.
The calculations are straightforward but lengthy,
see \texttt{\seqsplit{com/polytechnik/kgo/LagrangeMultipliersNaiveKraus.java:getLagrangeMultipliers}}
for an implementation that,
given $b_{s,jk}$ and $S_{jk;j^{\prime}k^{\prime}}$,
calculates the Lagrange multiplier matrices $\lambda_{ij}$ and $\nu_{ss^{\prime}}$.

This approach differs from the commonly used one in that
we only utilize the Lagrangian variation (\ref{variationKrausLagrangiznZero})
to calculate Lagrange multipliers.
The $b_{s,jk}$
is then determined from the eigenproblem solution with these Lagrange multipliers
used in (\ref{KrausSwithLagrange}).
This allows us
to apply constraint satisfying adjustments to the $b_{s,jk}$
before using it in the calculation of Lagrange multipliers.
We cannot simultaneously solve the optimization problem for 
Lagrange multipliers $\lambda_{ij}$, $\nu_{ss^{\prime}}$
and the solution $b_{s,jk}$. Instead, on every iteration,
we adjust the $b_{s,jk}$ to satisfy all required constraints,
solve the linear system to find the Lagrange multipliers,
and then use them in an eigenproblem to find the next $b_{s,jk}$.

\subsection{\label{LinearConstraints}Convergence-Helper Linear Constraints}
The problem we study involves quadratic constraints
(\ref{KrausOrt1}) and (\ref{KrausCanonic}).
These constraints cannot be directly incorporated into the eigenproblem
(\ref{KrausOptimizationVarS}).
However, homogeneous linear constraints can be easily incorporated.
Let's construct the linear constraints corresponding to the constrained local variation of the current iteration $b_{s,jk}$.
The concept can be illustrated by varying a quadratic constraint 
$\Braket{\mathbf{x}|\mathbf{y}}=0$ to obtain
a linear constraint
 on $\delta\mathbf{x}$ and $\delta\mathbf{y}$
at fixed $\mathbf{x}=u_{j*}$ and $\mathbf{y}=u_{j^{\prime}*}$:
$\delta\Braket{\mathbf{x}|\mathbf{y}}=\Braket{\delta\mathbf{x}|\mathbf{y}}
+\Braket{\mathbf{x}|\delta\mathbf{y}}=0$;
similarly, two expressions $1=\Braket{\mathbf{x}|\mathbf{x}}=\Braket{\mathbf{y}|\mathbf{y}}$
are replaced by 
 $\Braket{\mathbf{x}|\mathbf{x}}-\Braket{\mathbf{y}|\mathbf{y}}=0$.
By varying it, obtain a linear constraint on
 $\delta\mathbf{x}$ and $\delta\mathbf{y}$:
$\Braket{\delta\mathbf{x}|\mathbf{x}}+\Braket{\mathbf{x}|\delta\mathbf{x}}
-\Braket{\delta\mathbf{y}|\mathbf{y}}-\Braket{\mathbf{y}|\delta\mathbf{y}}=0$.
In the case of unitary mapping ($D = n$, $N_s = 1$), this corresponds to replacing the $n(n+1)/2$
quadratic constraints of the unitary property (\ref{FUMaxConstraints}) with $n(n+1)/2 - 1$
homogeneous linear constraints (\ref{KrauslinearConstraintsHomog}), that simply reduce the search space dimension  (\ref{KrausuExprVp}),
and a single quadratic constraint (\ref{KrausPartialConstraint}),
referred to as a simplified (partial) constraint.
This problem can then be iteratively solved with excellent convergence
using an eigenvalue problem solver as a building block.

Consider
(\ref{KrausOrt1}) $j\ne j^{\prime}$ off-diagonal elements. There are $D(D-1)/2$
total distinct ones. Variating (\ref{KrausOrt1}), we obtain
\begin{subequations}
 \label{offdiagC}
\begin{align}
C_{d;s,jk}&=b_{s,j^{\prime}k} \\
C_{d;s,j^{\prime}k}&=b_{s,jk}
\end{align}
\end{subequations}
Two equations
set different elements in $C_{d;s,jk}$ for the same $d$,
they may be viewed as two  initialization commands for the matrix $C_{d;s,jk}$.
In (\ref{offdiagC}), the constraint index $d$ takes  $D(D-1)/2$ distinct values
corresponding to all $j^{\prime}<j$ pairs.

Consider 
the inhomogeneous constraints (\ref{KrausOrt1})
corresponding to the diagonal elements with $j=j^{\prime}$.
There are $D$ of them. Since the partial constraint (\ref{KrausPartialConstraint}) preserves the total norm,
it is sufficient for all diagonal elements to be equal.
Equality of diagonal elements constitutes a homogeneous constraint, yielding 
$D-1$ constraints for a given $b_{s,jk}$.
\begin{subequations}
\label{diagC}
\begin{align}
C_{d;s,jk}&=b_{s,jk} \\
C_{d;s,j-1\,k}&=-b_{s,j-1\,k}
\end{align}
\end{subequations}
Similarly to the previous case, two equations set different elements in 
$C_{d;s,jk}$ for the same $d$.
In (\ref{diagC}) the constraint index $d$ takes $D-1$ distinct values
corresponding to $j=1\dots D-1$.
For an implementation of (\ref{offdiagC}) and (\ref{diagC})
 see
\texttt{\seqsplit{com/polytechnik/kgo/LinearConstraintsKraus.java:getOrthogonalOffdiag0DiagEq}}.

Consider the canonical form constraints on Kraus operators (\ref{KrausCanonic}).
There are no diagonal elements,
resulting in a total of $N_s(N_s-1)/2$ distinct constraints.
Similarly obtain
\begin{subequations}
 \label{offdiagCKraus}
\begin{align}
C_{d;s,jk}&=b_{s^{\prime},jk} \\
C_{d;s^{\prime},jk}&=b_{s,jk}
\end{align}
\end{subequations}
Two equations
set different elements in $C_{d;s,jk}$ for the same $d$,
they may be viewed as two  initialization commands for the matrix $C_{d;s,jk}$.
In (\ref{offdiagCKraus}) constraint index $d$ takes  $N_s(N_s-1)/2$ distinct values
corresponding to all $s^{\prime}<s$ pairs;
for $N_s=1$ the constraints vanish.
For an implementation of (\ref{offdiagCKraus})
see
\texttt{\seqsplit{com/polytechnik/kgo/LinearConstraintsKraus.java:getOrthogonalKrausOffdiag0}}.

Quadratic constraints for trace preservation (\ref{TracePreservingConstraint}) can generate $(n-1)(n+2)/2$
homogeneous linear constraints,
similar to how (\ref{offdiagC}) and (\ref{diagC})
were derived from the orthogonality constraint (\ref{KrausOrt1}).
\begin{subequations}
 \label{TracePreserveCKK}
\begin{align}
C_{d;s,jk}&=b_{s,jk^{\prime}} \\
C_{d;s,jk^{\prime}}&=b_{s,jk}
\end{align}
and
\begin{align}
C_{d;s,jk}&=b_{s,jk} \\
C_{d;s,jk-1}&=-b_{s,jk-1}
\end{align}
\end{subequations}
see \texttt{\seqsplit{com/polytechnik/kgo/LinearConstraintsKraus.java:getTracePreservingConstraints}} for an implementation.

These linear homogeneous convergence-helper constraints can be summarized as follows:
\begin{itemize}
\item (\ref{offdiagC}) --- $D(D-1)/2$ in total, corresponding 
to (\ref{KrausOrt1}), ensures that off-diagonal elements are zero.
\item (\ref{diagC}) --- $D-1$ in total, corresponding to (\ref{KrausOrt1}),
ensures that the diagonal elements are equal.
\item (\ref{offdiagCKraus}) --- $N_s(N_s-1)/2$ in total, corresponding to (\ref{KrausCanonic}).
\item (\ref{TracePreserveCKK}) --- $(n-1)(n+2)/2$ in total,
corresponding to (\ref{TracePreservingConstraint}), optionally used
instead of (\ref{offdiagC}) and (\ref{diagC}).
\end{itemize}
The combined set of linear constraints can be put to (\ref{KrauslinearConstraintsHomog})
with $N_d=(D-1)(D+2)/2+N_s(N_s-1)/2$
to restrict\footnote{
\label{optionalConstraintsTracePreserve}
From the
$n(n+1)/2$
trace preservation constraints (\ref{TracePreservingConstraint}),
we obtained
$(n-1)(n+2)/2$
linear constraints (\ref{TracePreserveCKK}).
These can be used instead of (\ref{offdiagC}) and (\ref{diagC}).
For $n=D$
both sets are equivalent.
When, for some reason, they are used together, it is possible to have redundant constraints
(e.g., in the $D=n$ unitary case where $N_s=1$).
In such cases Gaussian elimination retains only the $\mathrm{rank}\,C_{d;s,jk}$
of them.} the variation subspace of
$b_{s,jk}$.
All $C_{d;s,jk}$
are calculated from $b_{s,jk}$,
similar to the calculation of 
Lagrange multipliers in
Appendix \ref{lagrangeMultipliersCalculation}.
Our algorithm uses, as an iteration state,
not a pair: approximation, Lagrange multipliers $(b_{s,jk},\{\lambda_{ij},\nu_{ss^{\prime}}\})$,
but a triple: approximation, Lagrange multipliers, homogeneous linear constraints:
$(b_{s,jk},\{\lambda_{ij},\nu_{ss^{\prime}}\},C_{d;s,jk})$;
it is the linear constraints themselves that ensure algorithm convergence
 ---
current work and previous \cite{belov2024partiallyPRE}
results show their critical importance.

\section{\label{SchodingerNonStationary}A Time-Dependent Schr\"{o}dinger-like Equation}
A formulated novel algebraic problem (\ref{eigenvaluesLikeProblem})
is a generalization of the eigenvalue problem (time-independent Schr\"{o}dinger equation).
A question arises regarding the generalization of (\ref{eigenvaluesLikeProblem}) to a time-dependent form.
Consider the equation
\begin{align}
i\hbar \frac{\partial \mathcal{U}}{\partial t}&=S \mathcal{U}
\label{SchodingerNonStationaryEq}
\end{align}
where superoperator $S$ is a Hermitian tensor
$S_{jk;j^{\prime}k^{\prime}}$
and 
$\mathcal{U}$ is a unitary operator $u_{jk}$ where $D=n$.
This equation is different from the dynamic equation for the density matrix
\begin{align}
i\hbar\frac{\partial \rho}{\partial t}=\mathcal{L} \rho
\label{rhoDynamics}
\end{align}
where the Liouville operator $\mathcal{L} \rho = H\rho - \rho H$.
This equation has the Hermitian density matrix $\rho$ as its solution.
The distinction arises from the fact that the operators $\mathcal{L}$ and $S$ have completely different structures,
and the solution $\mathcal{U}$ is now a unitary operator,
not a Hermitian matrix $\rho$.
We can, however, employ a transition from the wavefunction Schr\"{o}dinger equation to
density matrix
\href{https://en.wikipedia.org/wiki/Density_matrix#The_von_Neumann_equation_for_time_evolution}{Liouville -- von Neumann}
equation (\ref{rhoDynamics})
to construct an equation for the density supermatrix $\Upsilon$, which represents the density tensor.
The density matrix is a convex combination of pure states (Eq. \ref{densMatr}); this results in a transition from a vector $\Ket{\psi}$ of dimension $n$ to a Hermitian matrix $\rho$ of dimension $n \times n$.
Similarly, consider a convex combination of unitary channels
\begin{align}
\Upsilon&=\sum\limits_{s=0}^{Dn-1} P^{[s]} \Ket{\mathcal{U}^{[s]}} \Bra{\mathcal{U}^{[s]}}
\label{densMatrConvexSuperpos}
\end{align}
This results in a transition from a $D \times n$ unitary matrix $\mathcal{U}$ (where $D = n$) to a $Dn \times Dn$ density tensor $\Upsilon$,
which has the same structure as the Hermitian tensor $S_{jk;j^{\prime}k^{\prime}}$.
A general quantum channel is described by a density tensor $\Upsilon$.
Then, similarly to Eq. (\ref{rhoDynamics}), we can assume that $\Upsilon$ satisfies its own Liouville equation
\begin{align}
i\hbar\frac{\partial \Upsilon}{\partial t}=\mathcal{L} \Upsilon
\label{YDynamics}
\end{align}
Equation (\ref{YDynamics}) is analogous to Eq. (\ref{rhoDynamics}), but describes dynamics
where a unitary operator $\mathcal{U}$ replaces the vector wavefunction $\Ket{\psi}$,
and the density tensor $\Upsilon$ replaces the density matrix $\rho$.
The specific form of the operator $\mathcal{L}$ requires further research; the first form to consider is, evidently,
\begin{align}
\mathcal{L} \Upsilon &= S \Upsilon - \Upsilon S
\label{SU}
\end{align}
The commutator is possible because $S$ and $\Upsilon$ have the same tensor structure.
This form, however, differs from regular density matrix dynamics.
For a pure superstate
$\Upsilon = \Ket{\mathcal{U}} \Bra{\mathcal{U}}$,
where $\mathcal{U}$ is the solution of (\ref{eigenvaluesLikeProblem}),
$\mathcal{L} \Upsilon$ from (\ref{SU}) is not zero:
$(\mathcal{L} \Upsilon)_{ik;jq}=
\sum_{j^{\prime}=0}^{D-1}\lambda_{ij^{\prime}}u_{j^{\prime}k}u^*_{jq}
-
\sum_{j^{\prime}=0}^{D-1}u_{ik}u^*_{j^{\prime}q}\lambda^*_{jj^{\prime}}
$.
This issue does not arise in regular density matrix dynamics,
where for a Hamiltonian eigenvector $\Ket{\psi}$,
the density matrix $\rho = \Ket{\psi}\Bra{\psi}$ satisfies $0 = H\rho - \rho H$.

If the tensor $\Upsilon_{jk;j^{\prime}k^{\prime}}$ is known only in matrix form,
the expansion (\ref{densMatrConvexSuperpos}) can be obtained by applying the hierarchy construction
from Section \ref{Kraushierarchy} to $\Upsilon_{jk;j^{\prime}k^{\prime}}$.
Note that various orthogonality constraints can be applied to the operators
$\mathcal{U}^{[s]}$
in (\ref{densMatrConvexSuperpos}).
For example, the constraint (\ref{constraintsUPreviousDeniminator})
can be used instead of (\ref{constraintsUPreviousOrthogonalChannel}),
which was previously employed in Section \ref{Kraushierarchy}.
The operators $\mathcal{U}^{[s]}$ in (\ref{densMatrConvexSuperpos}) then satisfy the ``denominator''-type
orthogonality condition (\ref{constraintsUPreviousDeniminator})
\begin{align}
D\delta_{ss^{\prime}}&=
\Braket{\mathcal{U}^{[s]}|\mathcal{U}^{[s^{\prime}]}}
\label{constraintsUPreviousDeniminatorSuperDensity}
\end{align}
We will defer the study of the equation for the density tensor $\Upsilon$
to future research and focus here on a simple example of the ground state time evolution of Eq. (\ref{SchodingerNonStationaryEq}).
This equation is a generalization of the Schr\"{o}dinger equation from wavefunction vector space to the space of unitary operators.
The Schr\"{o}dinger equation itself can be derived from Brownian motion, Fisher information,
the Hamilton-Jacobi equation, and various other approaches to direct problems
(see references [21-34] in \cite{efthimiades2024derivation}).
Our equation (\ref{SchodingerNonStationaryEq}) could likely be derived from various inverse problems.

Consider a ground state stationary solution of (\ref{SchodingerNonStationaryEq}) corresponding to $\mathcal{U}^{[0]}$,
i.e. obtained by maximizing the fidelity $\mathcal{F}$ without using the constraints from (\ref{constraintsUPreviousOrthogonalChannel}).
This solution, represented by the matrix $u^{[0]}_{jk}$, satisfies (\ref{eigenvaluesLikeProblem}) with some $\lambda^{[0]}_{ij}$.
The matrix $\lambda^{[0]}_{ij}$ is Hermitian but not necessarily diagonal.
Let us convert the problem to the basis in which $\lambda^{[0]}_{ij}$ is diagonal.
Consider the basis of $\lambda_{ij}$ eigenvectors
\begin{align}
\sum\limits_{j=0}^{D-1}\lambda_{ij} \beta^{[p]}_j&=\lambda^{[p]} \beta^{[p]}_i
\label{evLambdaEigenproblem}
\end{align}
Then, if we formulate the original problem (\ref{eigenvaluesLikeProblem}) in basis
\begin{align}
v_{pk}&=\sum\limits_{j=0}^{D-1}\beta^{[p]}_ju_{jk}
\label{vTransformBasis}
\end{align}
the obtained solution is $v^{[0]}_{pk}$, and the corresponding $\lambda_{pp^{\prime}}$ is diagonal
with the diagonal elements equal to the eigenvalues $\lambda^{[p]}$ in (\ref{evLambdaEigenproblem}).
We considered a similar basis transformation in \cite{malyshkin2022machine}
(see Appendix A.2 therein, ``On Iteration Step Without Using the SVD''),
where $S_{jk;j^{\prime}k^{\prime}}$ and $u_{jk}$ are converted between bases
for the purpose of improving the convergence of an algorithm.
However, later \cite{belov2024partiallyPRE} we found an algorithm that ensures convergence in any basis.
Here, this transformation is performed solely to obtain the Lagrange multipliers
$\lambda_{ij}$ (\ref{newLambdaSolPartial}) in diagonal form,
which is necessary to find a stationary state solution of the time-dependent equation (\ref{SchodingerNonStationaryEq}).
Thus, without loss of generality, the $\lambda_{ij}$ can be considered diagonal;
otherwise, the original problem basis should be changed to (\ref{vTransformBasis}).

Let us also rewrite (\ref{SchodingerNonStationaryEq}) by
replacing
$S$ with $\lambda_{jj^{\prime}}\delta_{kk^{\prime}}$,
and referring to this $S$ as the single Hamiltonian approximation.
\begin{align}
S_{jk;j^{\prime}k^{\prime}}&\approx\lambda_{jj^{\prime}}\delta_{kk^{\prime}}
\label{groundStateApproximation}
\end{align}
In this approximation, a general $S$, which encompasses many different Hamiltonians,
is replaced by a form containing only a single Hamiltonian $\lambda$,
typically corresponding to the ground state.
The maximal fidelity
$\mathcal{F}=\Braket{\mathcal{U}|\lambda|\mathcal{U}}=\mathrm{Tr} \lambda$
then holds for an arbitrary unitary operator $\mathcal{U}$,
not just for the ground state solution of the equation (\ref{eigenvaluesLikeProblem}).
By focusing solely on the ground state quantum channel
the dynamic equation (\ref{SchodingerNonStationaryEq}) becomes
\begin{align}
i\hbar \frac{\partial \mathcal{U}}{\partial t}&=\lambda \mathcal{U}
\label{SchodingerNonStationaryEqMatrProduct}
\end{align}
This equation is akin to a Schr\"{o}dinger equation with the Hamiltonian $\lambda$,
where all possible solutions are encompassed in $\mathcal{U}$.
The equation describes the simultaneous time evolution of all possible solutions.
The full superoperator $S_{jk;j^{\prime}k^{\prime}}$ contains many Hamiltonians.
For each solution of (\ref{eigenvaluesLikeProblem}), the Lagrange multiplier matrix $\lambda$
can be considered as a Hamiltonian of some quantum system.
For $u_{pk}$ being the ground state solution $u^{[0]}_{pk}$ of (\ref{eigenvaluesLikeProblem})
and $\lambda_{pj}$ being diagonal $\lambda^{[p]}\delta_{pj}$,
the time-dependent solution of it is
\begin{align}
u_{pk}(t)&=\exp\left(-\frac{i}{\hbar}\lambda^{[p]}t\right)u_{pk}(t=0)
\label{nonStationarySolt}
\end{align}
This can be viewed as $u_{p*}$ vectors evolving each with its own phase $\exp\left(-it\lambda^{[p]}/\hbar\right)$.
If we consider the operator $u_{pk}$ as $D$ ``wavefunctions'' $u_{p*}$ of dimension $n$,
then each of them evolves with its own ``energy'' $\lambda^{[p]}$.
This represents the time evolution of the ground state solution of (\ref{eigenvaluesLikeProblem}).
The difference from traditional quantum mechanics is that the ground state $u_{pk}$
now contains multiple vectors, each evolving with its own exponent $\lambda^{[p]}$.
The density tensor corresponding to the ground state is
\begin{align}
&\left(\Ket{\mathcal{U}} \Bra{\mathcal{U}}\right)_{pk;p^{\prime}k^{\prime}}(t)= \label{UpstPure} \\
&\quad\exp\left(-\frac{i}{\hbar}\left(\lambda^{[p]}-\lambda^{[p^{\prime}]}\right)t\right)
u_{pk}(t=0)
u^*_{p^{\prime}k^{\prime}}(t=0)
\nonumber
\end{align}
Unlike the density matrix of a traditional quantum system's ground state, which does not depend on time,
this density tensor has only the diagonal elements, $p = p^{\prime}$, that are independent of time.
This explains why (\ref{SU}) is nonzero for a pure superstate $\Upsilon = \Ket{\mathcal{U}} \Bra{\mathcal{U}}$,
where $\mathcal{U}$ is the solution to (\ref{eigenvaluesLikeProblem}).

A \href{https://en.wikipedia.org/wiki/Heat_equation#Statement_of_the_equation}{heat transfer}-like
equation can be obtained by removing the imaginary unit $i$ from (\ref{SchodingerNonStationaryEq}),
resulting in $\varkappa \partial \mathcal{U}/\partial t=S \mathcal{U}$.
In this case, the solution (\ref{nonStationarySolt}) results in an exponentially growing or decaying $u_{pk}(t)$.

The time evolution of a superposition of two solutions
is more complicated compared to the Schr\"{o}dinger   equation.
For the Schr\"{o}dinger  equation, a superposition of two eigenstates $a\psi^{[0]} + b\psi^{[1]}$
of a Hamiltonian time-evolves as $a\psi^{[0]}\exp\left(-itE_0/\hbar\right) + b\psi^{[1]}\exp\left(-itE_1/\hbar\right)$.
At the same time, if we consider a superposition of two solutions
$au^{[0]}_{jk} + bu^{[1]}_{jk}$ 
of (\ref{eigenvaluesLikeProblem}), the result may not be unitary.
The violation of unitarity resulting from such a superposition has deep physical significance.
Such a superposition does not describe a physical state and, therefore, should not be considered.
A proper generalization is the introduction of the density supermatrix $\Upsilon$ (\ref{densMatrConvexSuperpos}),
which describes a mixed type of state.

Another feature that may differ between wavefunctions and quantum channels
is the state post-measurement destruction rules.
In traditional quantum mechanics, the state $\Ket{\psi}$ is destroyed (at least partially)
after the measurement act $\Braket{\psi|R|\psi}$.
In contrast, the measurement $\Braket{\mathcal{U}|R|\mathcal{U}}$ involves a quantum channel $\mathcal{U}$
as the state.
The post-measurement destruction rules for a quantum channel $\mathcal{U}$ remain an open question.
It is possible that these rules differ between wavefunctions and quantum channels.
A very important question is what the measurable unit is in quantum channel dynamics:
whether it is a scalar or a Hermitian matrix?
Since the corresponding algebraic problem (\ref{eigenvaluesLikeProblem}) has eigenmatrices as its spectrum,
rather than the usual eigenvalues (scalars),
we are inclined to believe that the measurement unit is a Hermitian matrix.
As the ground state energy of a quantum system can be obtained as a result of a single measurement act,
a  quantum channel single measurement could potentially provide the Hamiltonian $\lambda^{[0]}$
corresponding to the solution of (\ref{eigenvaluesLikeProblem}) with the maximal fidelity.

For these reasons, we will leave a detailed consideration of the time-dependent equations
(\ref{SchodingerNonStationaryEq}) and (\ref{YDynamics})
for future research and limit ourselves to the time evolution (\ref{nonStationarySolt}) of the ground state,
which corresponds to each $u_{p*}$ vector evolving with its own ``energy'' $\lambda^{[p]}$.
The total fidelity is equal to the sum of all $\lambda^{[p]}$, and the ground state is the state with the maximum total fidelity.
The ground state solution corresponds to a unitary operator $\mathcal{U}(t)$
that itself depends on time (\ref{nonStationarySolt}).

\section{\label{StatesWithMemory}Parametrizing states with memory using a feedback loop}

In this and previous works,
we considered a
\href{https://en.wikipedia.org/wiki/Quantum_channel#Memoryless_quantum_channel}{memoryless quantum channel}
described by a set of Kraus operators $B_s$ or, in the simplest case where $N_s=1$, by a single unitary operator $\mathcal{U}$.
A question arises about how memory can be introduced into the model.
Modeling a system with even a single bit of internal state is a daunting task,
see Appendix H of \cite{malyshkin2019radonnikodym}, where we discuss the problem of modeling a
\href{https://en.wikipedia.org/wiki/Flip-flop_(electronics)#Classical_positive-edge-triggered_D_flip-flop}{synchronous positive-edge-triggered D flip-flop}
(D trigger) with ML.
Any memory can be represented by a feedback loop,
as seen in, for example, a D-trigger or a
\href{https://en.wikipedia.org/wiki/Recurrent_neural_network}{recurrent neural network}.
The question is how to introduce memory into
the quantum channel (\ref{operatorTransform})
described by an operator $\mathcal{U}$?
For pure states, the channel transforms
an input state $\Ket{\psi}$ into the output state $\Ket{\mathcal{U}|\psi}$.
Let's assume there is a feedback loop that requires some (but not all)
components of the output state to be equal (within a phase)
to the same components of the input state.
\begin{align}
&\hbox {
\begin{circuitikz}
\ctikzset{multipoles/thickness=3}
\ctikzset{multipoles/dipchip/width=2}
\draw (0,0) node[dipchip,
num pins=8, no topmark,
external pins width=0.1,
hide numbers
,draw only pins={1,2,4,5,7,8}
](C){$\Ket{\psi}\to\Ket{\mathcal{U}|\psi}$};
\node [left, font=\tiny] at (C.pin 1) {$x_0$};
\node [left, font=\tiny] at (C.pin 2) {$x_1$};
\node [left, font=\tiny] at (C.pin 3) {$\bm{\vdots}$};
\node [left, font=\tiny] at (C.pin 4) {$x_{\{k\}}$};
\node [right, font=\tiny] at (C.pin 8) {$y_0$};
\node [right, font=\tiny] at (C.pin 7) {$y_1$};
\node [right, font=\tiny] at (C.pin 6) {$\bm{\vdots}$};
%\draw (C.pin 6) to[short,-, v=$\vdots$] ++ (0.1,0);
\node [right, font=\tiny] at (C.pin 5) {$y_{\{k\}}$};
\draw [blue,thick](C.pin 5)
to[short,-] ++(0.0,-0.8)
to[short,-,i=$\lambda^{[i]} x^{[i]}_{\{k\}}$] ++(-3,0) -| (C.pin 4)
;
\end{circuitikz}
} \nonumber \\
&\psi^{[i]}_{\{k\}}=\lambda^{[i]} \sum\limits_{k=0}^{n-1} u_{\{k\}k}\psi^{[i]}_k
\label{psiEVk}
\end{align}
Here the $\{k\}$ denotes these selected ``feedbacked'' components.
If $\{k\}$  is an empty set, we obtain the problem we considered in \cite{belov2024partiallyPRE}.
If $\{k\}$ includes \textsl{all} $n$ components ---
then (\ref{psiEVk}) becomes a regular eigenproblem with
$\lambda$ being $\exp(i\zeta)$ (or $\pm 1$ for an orthogonal $u_{jk}$).
In this case the problem does not have any input and the internal state can be represented as a superposition of $n$ eigenstates of eigenproblem (\ref{psiEVk}).
Now consider $\{k\}$ to be a subset of $0\dots n-1$
indices
with a total of $n_m$ elements.
Then the problem (\ref{psiEVk}) represents an eigenvalue-like
problem of dimension $n_m$, $i=0\dots n_m-1$,
where an ``eigenvector'' $\Ket{\psi^{[i]}}$ (there are $n_m$ total)
has $n-n_m$ components set externally
and $n_m$ components being an internal state.
This ``partial eigenvalue'' (\ref{psiEVk}) problem is a simple way
to represent a quantum channel with an internal state (memory).
A given state $\Ket{\psi}$ has $n-n_m$ free components (input)
and $n_m$ components of the internal state.
A change in the input components can potentially cause changes in the internal
state. The study of unitary quantum channels
with internal states is a subject for our future research.

A different application of these ``feedbacked'' states
of a unitary operator relates to the problem of integer factorization.
An integer can be encoded into a state $\Ket{\psi}$,
and the multiplication of the integer modulo $N$
by a number $a$ can be represented as a unitary transformation $\mathcal{U}$.
In Shor's algorithm \cite{shor1999polynomial},
a major step in the prime factorization of a number $N$ involves finding the period $r$ of
\begin{align}
a^{r}\equiv 1 {\bmod {N}}
\label{periodR}
\end{align}
If multiplication by $a$ is encoded as a unitary transformation $\mathcal{U}$
acting on the state $\Ket{\psi}$, then the problem (\ref{periodR})
can be framed as finding the $r$ such that $\Ket{\mathcal{U}^r | \psi_0}$
equals the initial state $\Ket{\psi_0}$ up to a phase factor.
This is similar to the concept of the feedback loop shown in (\ref{psiEVk}).
For example, if we were able to build an actual quantum system with the Hamiltonian
\begin{align}
H&=i\frac{\hbar}{\tau} \ln \mathcal{U}
\label{logUCalc}
\end{align}
and initial state $\Ket{\psi^{(t=0)}}=\Ket{\psi_0}$
then, since the system's time evolution
\begin{align}
  U&=
  \exp \left[-i\frac{t}{\hbar} H \right] \label{Uquantum} \\
   \Ket{\psi^{(t)}}&=\Ket{U \middle| \psi^{(t=0)}} \label{unitaryPsiEvolution}
\end{align}
the calculation of $r$ can be reduced to simply observing the quantum system
(\ref{logUCalc})
with $\mathcal{U}=U(t=\tau)$
at time moments
$t = \tau l$ and waiting until $\Ket{\psi^{(t=\tau l)}}$ matches $\Ket{\psi_0}$ up to a phase.
The index $l$ of this time moment gives the period $r$.
For this quantum system, the ``Boltzmann time'' of spontaneous return to the exact
initial configuration provides the sought period $r$.

\section{\label{compComplexityEstimation}An estimate of computational complexity}
The described class of algorithms solves QCQP problems that arise in the maximization of quadratic fidelity,
subject to quadratic constraints on mapping operators (e.g., unitarity).
This problem is equivalent to an algebraic problem; for example,
in the case of unitary learning, it corresponds to the algebraic problem (\ref{eigenvaluesLikeProblem})
originally introduced in \cite{malyshkin2019radonnikodym}.
This represents the simplest problem of this type.
In the general case, this is a new algebraic problem of dimension $N_sDn$.
It can be solved numerically using an iterative algorithm that, on each iteration, replaces $N_c$
quadratic constraints with $N_c-1$ homogeneous linear constraints
and a single quadratic constraint (\ref{KrausPartialConstraint}), referred to as a simplified (partial) constraint,
see Appendix \ref{LinearConstraints} above.
Since a QCQP problem with a single quadratic constraint is equivalent to an eigenvalue problem,
a regular eigenvalue problem solver is applied to a problem of dimension $N_V=N_sDn-N_c+1$,
this is step \ref{firstStepLambda} of the algorithm.
As discussed in Appendix B of \cite{belov2024partiallyPRE},
this is the most computationally intensive step.
Its computational complexity in the unitary learning case ($D=n$, $N_s=1$)
involves solving an eigenproblem with a matrix of dimension $N_V=n^2 - n(n+1)/2+1$.
This complexity can be estimated as $O(n^4)$ when using a specialized solver optimized
to find only the maximal eigenvalue and $O(n^6)$
when using a general-purpose solver.
Our current implementation \cite{polynomialcode} uses a general-purpose eigenvalue solver.

Let us compare the computational complexity of this ``algebraic'' algorithm with that of the
``mathematical analysis'' ones: second-order Newton's method and first-order gradient method.
A unitary matrix of dimension $n$ can be completely parametrized with $N=n(n+1)/2$ parameters.
There are $n$ fewer parameters if we do not need unmeasurable phases.
The number of parameters is always greater than or equal to $n(n-1)/2$ (e.g., Euler angles),
see (\ref{nonStationarySolt}), which gives a time-dependent ground state unitary operator $\mathcal{U}(t)$ such that,
at every time moment $t$, the matrix $\mathcal{U}$ gives the maximal value of fidelity $\mathcal{F}$ (\ref{FUMax}).
The complexity of the classical Newton's method for nonlinear systems of $N$
equations is $O(N_{it}N^3)$, as a linear system needs to be solved at each iteration.
Some tricks, such as updating the Jacobian only once every $m$ iterations,
can reduce the complexity to $O(N_{it}N^2+N_{it}N^3/m)$.
Since the number of iterations, $N_{it}$, for Newton's method is often independent of $N$,
the practical complexity of the Newton's method algorithm can be estimated
to lie between $O(n^4)$ and $O(n^6)$,\footnote{
If all $M$ observations are not combined into a single entity,
such as $S_{jk;j^{\prime}k^{\prime}}$,
then the complexity may range between $O(Mn^4)$ and $O(Mn^6)$, which significantly worsens the result.
}
depending on the Jacobian updating strategy.
While the computational complexity of the proposed algebraic algorithm is comparable to that of the second-order Newton's method,
using the generalized eigenvalue problem as the algorithm's building block offers the advantage
of obtaining multiple solution candidates (eigenvectors).
Numerical experiments\cite{belov2024partiallyPRE} demonstrate that this significantly
increases the chances of finding the global maximum.

The computational complexity of the gradient method in the general case is
$O(N_{it}MN)$.
If the $M$-sum can be factored out from the objective functional (e.g., into a form like (\ref{FUMax})),
the complexity reduces to $O(N_{it}N^2)=O(N_{it}n^4)$.
However, in the gradient method, the number of iterations $N_{it}$ is not guaranteed to be small,
nor is the algorithm guaranteed to converge,
especially for a global optimization problem that is nonconvex,
with local extrema and multiple saddle points.

The developed algorithm exhibits computational complexity on par with second-order
Newton's method and higher than that of the first-order gradient method.
Its main feature is the algebraic approach, which enables the construction of a globally converging algorithm.
While we lack a formal proof of its convergence, among millions of test runs,
only a few failed to converge to the global maximum.
In this work, we aimed to address a practical problem that can be validated using a classical computer.
Consider a scenario where we have input data (e.g., in the form of files on disk),
and we seek to recover the corresponding quantum mapping.
This problem frequently arises in Machine Learning for large dimensions,
as discussed in \cite{arjovsky2016unitary} and over 1,000 subsequent papers citing it.
In the notations of the submitted paper, typical dimensions of the ML problem are
$N_s = 1$ and $D = n \sim 10^3$ or greater.
The solution involves a form of unitary parametrization\cite{RazaUnitaryParametrization},
such as representing the unitary as a series of elementary rotations (e.g., Euler angles)
and employing a gradient-based optimization method.
The resulting solution is obtained through this process, but there is no guarantee that it corresponds to the global maximum.
Our primary focus was on developing an algorithm capable of providing the \textsl{global} maximum solution.
For this reason we considered an exact unitary mapping and aimed to recover it from a sample,
this is why we did not include noise or other common features in our study.
This is a highly challenging nonconvex problem, as the number of parameters grows quadratically with $n$ \cite{andersson2007finding}.
To the best of our knowledge, practical solutions are feasible only for small $n$ (fewer than a dozen)
or under specific setups \cite{schafers2024modified,ahmed2023gradient,wang2024variationalPRL}.
In our numerical experiments for unitary learning, the problem's dimensionality appears to be constrained solely by computational complexity.
At each iteration step, an eigenvalue problem of dimension $N_V=n^2 - n(n+1)/2+1$  must be solved,
with only the eigenstate corresponding to the largest eigenvalue being required.
Given the extensive global effort devoted to developing efficient numerical solvers for eigenproblems,
we anticipate that the dimensionality of the problem can be significantly increased.
Currently, however, we lack access to hardware capable of solving eigenproblems for matrices larger than dimension 1000.
Our original goal was to push the problem's dimensionality to a point where exact recovery
of orthogonal (real unitary) matrices would become infeasible.
However, computational limits were encountered first.
In our experiments, we tested the exact recovery of thousands of randomly generated orthogonal matrices
with dimensions under 50 (corresponding to an eigenproblem size of 1226).
The ability to find the global maximum remains the algorithm's most significant feature.

The computational complexity of constructing the tensor $S_{jk;j^{\prime}k^{\prime}}$ (\ref{SfunctionalClassic})
was insignificant for the pure state unitary mapping (\ref{mlproblemVector}) discussed in \cite{belov2024partiallyPRE}.
However, for the unitary mapping (\ref{operatorTransform}) of mixed states (\ref{mlproblemVectorDensityMatrix})
considered in the present work,
obtaining the fidelity in quadratic form requires transforming the mapping
to a density matrix square root mapping (\ref{mlproblemVectorDensityMatrixSQRT}).
This transformation may significantly increase the computational complexity
of calculating $S_{jk;j^{\prime}k^{\prime}}$ (\ref{StendorDensityMatrixSQRTrho}).
Using an eigenvalue problem to calculate the square root of a matrix would require solving
$M$ eigenproblems with the matrices $\rho$ and $\varrho$, each of dimension $n$,
resulting in a complexity of the order $O(Mn^3)$.
This cannot be reduced as in the QCQP case above, where only a single eigenvector was needed.
For the density matrix square root, we need all eigenvectors,
and the calculation of (\ref{StendorDensityMatrixSQRTrho}) probably cannot be done any better than $O(Mn^3)$.
The factor before that can, however, be reduced.
When the number of observations $M$ is large,
instead of using the regular method of calculating the matrix square root by solving an eigenproblem
(as in (\ref{KrausJJmSqrtG}) and (\ref{KrausKKmSqrtG}) above),
one can apply alternative methods to compute the square root of a positively definite Hermitian matrix without solving an eigenproblem \cite{bjorck2024numerical, higham1986newton}.
These approaches, like the eigenvalue problem, have a dominant complexity of $O(n^3)$
for the square root computation.
However, in practice, it can significantly improve computational efficiency when creating
the mapping (\ref{mlproblemVectorDensityMatrixSQRT}) from the original data (\ref{mlproblemVectorDensityMatrix}).
Despite the complexity of calculating $S_{jk;j^{\prime}k^{\prime}}$ for a large $M$,
this creates no difficulty in applications since the calculations are similar to those used,
for example, in covariance matrix computation.
Each component of the tensor is a sum over all $M$ observations, and this task can be trivially parallelized.

The general Kraus case, $N_s>1$, is significantly more difficult,
since the fidelity (\ref{FasOperatorsProduct}), as a quadratic form, can only be obtained as an approximation.
However, once the approximation is obtained,
in the general case, it requires solving an eigenproblem of dimension $N_V=N_sDn-N_c+1$
at each iteration (step \ref{firstStepLambda} of the algorithm),
with only a single eigenvector corresponding to the maximal eigenvalue being needed.
For a general quantum channel (\ref{KrausOperator}),
we were not able to construct a converging algorithm, not to mention the increased computational complexity
(eigenvalue problem dimension is increased in Kraus rank times).
However, if we limit ourselves to mixed-unitary channels (\ref{UnitaryHierarchyQC}),
then this type of quantum channel can be reconstructed as a hierarchy of unitary operators.
Computationally, the problem in Section \ref{Kraushierarchy} is much simpler
than a general quantum channel reconstruction.
Instead of dealing with a single complicated general quantum channel problem,
the problem of unitary mapping needs to be solved several times.
Numerically, we were able to obtain a hierarchy of no more than $7$ operators,
which is probably due to the suboptimal algorithm for constraint conflict resolution used in Appendix \ref{KrausAdjustOrthogonal}.
The result is a mixed unitary channel (\ref{UnitaryHierarchyQC}).
However, contrary to the $N_s=1$ case, for $N_s>1$,
this is only an approximation, since we were not able to represent the proper fidelity (\ref{fidelityStdDefinition})
as the quadratic form of Eq. (\ref{FasOperatorsProduct}) in Kraus operators $B_s$ when $N_s>1$.

\bibliography{LD,mla}

\end{document}